\documentclass[letterpaper]{article} % DO NOT CHANGE THIS
\usepackage{aaai23}  % DO NOT CHANGE THIS
\usepackage{times}  % DO NOT CHANGE THIS
\usepackage{helvet}  % DO NOT CHANGE THIS
\usepackage{courier}  % DO NOT CHANGE THIS
\usepackage[hyphens]{url}  % DO NOT CHANGE THIS
\usepackage{graphicx} % DO NOT CHANGE THIS
\usepackage{comment}
\usepackage{natbib}  % DO NOT CHANGE THIS AND DO NOT ADD ANY OPTIONS TO IT
\usepackage{caption} % DO NOT CHANGE THIS AND DO NOT ADD ANY OPTIONS TO IT
\usepackage{amsmath,amssymb}
\usepackage{subfig}
\usepackage{booktabs}
\usepackage{bibentry}
\usepackage{multirow}
\usepackage{xcolor}

\newtheorem{assump}{Assumption}

\newtheorem{thm}{Theorem}

\usepackage{array}

\newcolumntype{C}[1]{>{\centering\let\newline\\\arraybackslash\hspace{0pt}}m{#1}}

\renewcommand{\hat}{\widehat}
\def\shownotes{1} 
 \ifnum\shownotes=1
\newcommand{\authnote}[2]{{[#1: #2]}}
\else 
\newcommand{\authnote}[2]{{}}
\fi
\newtheorem{lemma}{Lemma}

  \newcommand{\gnorm}[1]{{\vert\kern-0.25ex\vert\kern-0.25ex\vert #1 
		\vert\kern-0.25ex\vert\kern-0.25ex\vert}}

\usepackage{newfloat}
\usepackage{listings}
\DeclareCaptionStyle{ruled}{labelfont=normalfont,labelsep=colon,strut=off} % DO NOT CHANGE THIS
\usepackage[ruled]{algorithm2e}
\usepackage{algpseudocode}
\title{Semantic-Enhanced Image Clustering}
\author{
    Shaotian Cai\equalcontrib,
    Liping Qiu\equalcontrib,
    Xiaojun Chen~\footnote{Corresponding author.},
    Qin Zhang,
    Longteng Chen
}
\affiliations{
    \textsuperscript{\rm 1}Shenzhen University, Shenzhen, China\\
    cai.st@foxmail.com, qiuliping2021@email.szu.edu.cn, \{xjchen, qinzhang\}@szu.edu.cn, chenlt2021@163.com
}

\usepackage{bibentry}
\begin{document}

\maketitle

\begin{abstract}
Image clustering is an important and open-challenging task in computer vision. Although many methods have been proposed to solve the image clustering task, they only explore images and uncover clusters according to the image features, thus being unable to distinguish visually similar but semantically different images. In this paper, we propose to investigate the task of image clustering with the help of a visual-language pre-training model. Different from the zero-shot setting, in which the class names are known, we only know the number of clusters in this setting. Therefore, how to map images to a proper semantic space and how to cluster images from both image and semantic spaces are two key problems. To solve the above problems, we propose a novel image clustering method guided by the visual-language pre-training model CLIP, named \textbf{Semantic-Enhanced Image Clustering (SIC)}. In this new method, we propose a method to map the given images to a proper semantic space first and efficient methods to generate pseudo-labels according to the relationships between images and semantics. Finally, we propose performing clustering with consistency learning in both image space and semantic space, in a self-supervised learning fashion. The theoretical result of convergence analysis shows that our proposed method can converge at a sublinear speed. Theoretical analysis of expectation risk also shows that we can reduce the expected risk by improving neighborhood consistency, increasing prediction confidence, or reducing neighborhood imbalance. Experimental results on five benchmark datasets clearly show the superiority of our new method.

\end{abstract}

\section{Introduction}

Image classification, which assigns an image to a predefined set of classes, is an important task in computer vision. However, it is costly to obtain labeled data in the age of big data. To liberate us from laborious and trivial data labeling work, image clustering that aims to group images into different clusters without ground-truth semantic labels has become a more and more important task.

 \begin{figure}[!htb]
     \centering
     \includegraphics[width=.8\linewidth]{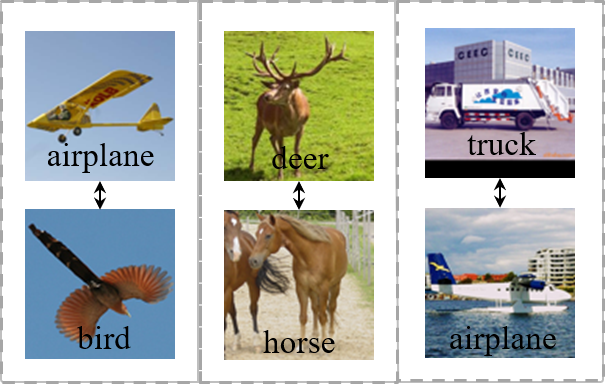}
     \caption{Visually similar but semantically different images on STL10 dataset, where every two images in a column are similar with the image embedding obtained by the CLIP pre-training model and the text in an image is its true label. }    
     \label{fig:visually_similar}
 \end{figure}

The early works in deep image clustering usually combine auto-encoders (AE) or Convolutional Neural Network (CNN) based representation learning with traditional shallow clustering methods~\cite{xie2016unsupervised,yang2016Joint, Yang2017Towards,Tian2017Deep,shaham2018spectral}. In recent years, with the rapid development of pre-training models, such as VGG-16~\cite{simonyan2014very}, Resnet~\cite{he2016deep}, ViT~\cite{dosovitskiy2020image}, Swin Transformer~\cite{liu2021swin}, image clustering methods leave the images representation task for pre-training model, and directly map image representations into labels by training a classification model like Multilayer Perceptron (MLP), by maximizing the mutual information between the image and its augmentations~\cite{ji2019invariant,li2021contrastive,zhong2021graph} or the likelihood of the cluster assignments between the image and its neighbors~\cite{wu2019deep,gansbeke2020scan,zhong2021graph,dang2021nearest}. However, since we want to obtain semantically meaningful clusters, it is difficult to solve this problem by only exploring images. Figure~\ref{fig:visually_similar} shows some examples 
that are semantically different but visually similar. For example, an image with an airplane may be visually similar to an image with a bird, and an image with a deer may be visually similar to an image with a horse.

\begin{comment}
  \begin{figure}[!htb]
     \centering
     \includegraphics[width=.8\linewidth]{diagram.png}
     \caption{Some examples of visually similar but semantically different subjects on STL10 dataset. We compute $k$ nearest neighbors for each image based on the embeddings from CLIP pre-training model to measure the visual similarities.}
     \label{fig:diagram_contrastive}
 \end{figure}
\end{comment}

\begin{figure}[t]
     \centering
     \includegraphics[width=1.0\linewidth]{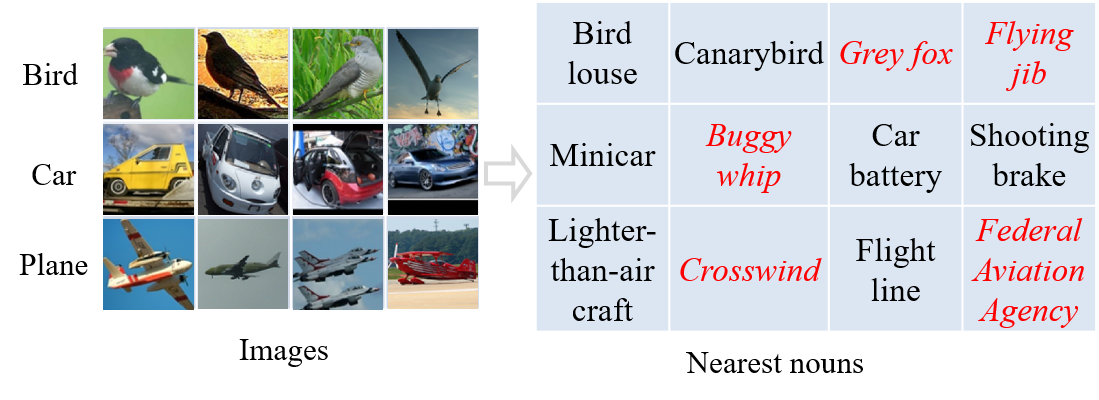}
     \caption{Images and their nearest nouns selected from WordNet~\cite{miller1995wordnet} on STL10, where the image and text embeddings are obtained via CLIP~\cite{radford2021learning}. The images corresponding to red italic nouns are wrongly mapped.}
     \label{fig:motivation}
 \end{figure}

Intuitively, we need to access the language model to improve image clustering with semantic information. Some works~\cite{jin2015cross,wang2020icmsc,yang2021jecl} try to explore the image-caption pairs to cluster images, but constructing the images with qualified captions is cost-intensive in real applications. Note the great success of visual-language pre-training models such as CLIP~\cite{radford2021learning}, which is trained on a dataset of 400 million image-text pairs available on the internet to align texts and images in common feature space by capturing the image-text relationships. It has shown surprising results in zero-shot learning tasks. However, we need to know the class names in zero-shot tasks, which hinders some potential applications such as image clustering when the class names are unavailable. This motivates us to utilize the visual-language pre-training model to compensate for the semantic information for better image clustering.

% \begin{figure}[!htb]
% \centering
%     \subfloat[\footnotesize ]{
%           \centering 
%           \label{fig:visually_similar}  
%           \includegraphics[width = .35\linewidth]{figures/visual.png}}\hspace{1mm}
%     \subfloat[\footnotesize ]{
%             \centering 
%           \label{fig:diagram_contrastive}
%             \includegraphics[width = .55\linewidth]{figures/motivation2.png}}
% \caption{(a) Some examples of visually similar but semantically different subjects on STL10 dataset. We compute $k$ nearest neighbors for each image based on the embeddings from CLIP pre-training model to measure the visual similarities. (b) An illustration of image clustering diagram and cross-modal guided image clustering diagram, where image clustering learns only from image information, while cross-modal guided image clustering learns both from image and text ones.}
% \label{fig:motivation2}
% \end{figure}

Although visual-language pre-training models such as CLIP can map images and texts into a unified space, Figure~\ref{fig:motivation} shows that simply mapping images to the nearest semantics does not improve the clustering. Therefore, in the task of image clustering with help of a visual-language pre-training model, we need to solve two key problems:
\begin{enumerate}
    \item \emph{How to map images to a proper semantic space that can improve the clustering?}
    \item \emph{How to cluster images from both image and semantic spaces?}
\end{enumerate}

In this paper, as shown in Figure~\ref{fig:framework}, we propose a novel image clustering method guided by the visual-language pre-training model CLIP, named \textbf{Semantic-Enhanced Image Clustering (SIC)}. The new method first maps the given images to a proper semantic space, generates pseudo-labels by taking the relationships between images and semantics into consideration, and then performs image clustering with consistency learning in both image space and semantic space. Our main contributions are summarized as follows:
\begin{itemize}
    \item We propose a method to select proper nouns to construct semantic space, and three methods to map images to semantics in order to generate pseudo-labels.
    \item The theoretical result on convergence shows that our proposed method can converge at a sublinear speed.
    \item The theoretical result on expectation risk shows that we can reduce the expected risk of our method by improving neighborhood consistency, increasing prediction confidence, or reducing neighborhood imbalance such that a sample lies in less sample's nearest neighborhoods.
    \item Experimental results on five benchmark datasets clearly show that SIC is superior to $20$ state-of-the-art and zero-shot learning with CLIP.
\end{itemize}

\begin{comment}
The current computer vision approaches exist several major problems, including labor intensive and high cost to create datasets, weak task adaption ability and so on. 

In the past, image clustering only considered the information of the image itself, and the image data was usually accompanied by some other information, such as tags and caption, etc. Recently, some Visual-Language models have constructed input data by collecting image and text information from the Internet, and combined them with representation learning to learn the relationship between image and text. This multimodal information is accompanied by higher levels of abstraction, providing the potential for making useful connections between different modals. The above findings motivate us to apply Visual-Language pre-training model to guide image clustering learning.

We map images to the corresponding semantic concepts in text space through the CLIP model. As shown in the Figure~\ref{fig:image2concept}, it is clearly observed that the same class of image will have some differences in semantic concept, so we can’t directly find a best performance concept for all the images belong to same class. It also means that the CLIP model more concentrates on text-level representation and unable to be directly used in the coarse-grained-level task like image clustering. This motivates us to search a more robust concept for images.
\end{comment}

\section{Related Work}

\subsection{Vision-Language Pre-training Models}
Vision-Language Pre-training (VLP) models align multi-modal data in common feature space by different pre-training tasks, which can be categorized into two categories: 1) VisualBert~\cite{li2019visualbert}, UNITER~\cite{chen2020uniter} and DALL-E~\cite{ramesh2021zero} use Language-based training strategy, including mask LM (Mask Language Modeling) such as Masked Language/Region Modeling, or autoregressive LM such as image caption and text-grounded image generation. 2) UNIMO~\cite{li2020unimo}, CLIP~\cite{radford2021learning}, ALIGN~\cite{jia2021scaling} utilize cross-modal contrastive learning to align the visual and textual information into a unified semantic space.

The core task of VLP is to model the interactions between images and texts, and there are two types of architectures for this: 1) The single-stream models like ImageBERT~\cite{qi2020imagebert}, Unicoder-VL~\cite{li2020unicoder} concatenate patch-wise or regional visual and textual embeddings and feed them to one encoder. 2) The dual-stream models like ViLBERT~\cite{lu2019vilbert} and CLIP~\cite{radford2021learning} obtain visual and textual embeddings with separate encoders.

Since VLP captures the relationships among images and texts (low-level semantics), in this paper, 
we propose to utilize the visual-language pre-training model to compensate for the semantic information for better image clustering.

\subsection{Image Clustering}

The early works in deep clustering usually simply combined feature learning with shallow clustering. For example, some methods combined the stacked auto-encoders (SAE) with the traditional clustering algorithms such as $k$-means~\cite{xie2016unsupervised,Yang2017Towards,Tian2017Deep}, subspace clustering~\cite{ji2017deep} and spectral clustering~\cite{shaham2018spectral}, or combined the Convolutional Neural Network (CNN) with the hierarchical clustering~\cite{yang2016Joint}. However, the above methods usually require post-processing to obtain cluster assignments.

Recently, some methods were developed to directly map images into labels with a classification model, by maximizing the mutual information between the labels of the original images and their augmentations~\cite{xu2019invariant,li2021contrastive,zhong2021graph}, or maximizing the likelihood of the cluster assignments between a sample and its nearest neighbors 
~\cite{zhong2021graph,dang2021nearest,chang2017deep,wu2019deep,gansbeke2020scan}. Some of them further generate pseudo-labels to refine the model~\cite{wu2019deep,gansbeke2020scan}. Furthermore, some methods were proposed to act as add-on modules to revise the classification model via label cleansing and retraining with the refined labels~\cite{gupta2020unsupervised,park2021improving}.

The pseudo-labels in~\cite{wu2019deep,gansbeke2020scan} are generated from the clustering results, and thus are doubtful. \cite{mahon2021selective} generates multiple groups of pseudo-labels by training multiple clustering algorithms independently, and sets the common pseudo-labels as high-quality pseudo-labels. However, it is cost-intensive, and the Hungarian algorithm makes it difficult to effectively align multiple groups of pseudo-labels.

In this paper, we propose to generate high-quality pseudo-labels according to the interaction between image and text by utilizing the vision-language model CLIP.

\section{Notation and Problem Definition}
In this paper, matrices are written as bold uppercase letters like $\mathbf{A}$. $\mathbf{a}_i$ represents the $i$-th row of $\mathbf{A}$, $a_{ij}$ represents the $i$-th row and the $j$-th column element of $\mathbf{A}$ and $\mathbf{A}^T$ is the transpose of $\mathbf{A}$. $\|\cdot\|_{1}$ expresses the $l_1$-norm of a vector. $\|\cdot\|$ donates the module of vector. 

Suppose we have an image dataset with $n$ instances sampled i.i.d. from input space $\mathcal{X} $ is denoted as $\mathcal{D}= \{x_1, x_2, \dots, x_n\}$, we can obtain the embeddings of these images as $\mathcal{U} = \{\mathbf{u}_1, \mathbf{u}_2, \dots, \mathbf{u}_n\}$ where $\mathbf{u}_i=g(x_i)$ is obtained via the image encoder $g(.)$ of CLIP. To capture the semantic meaning of these images, we introduce a semantic dataset $\mathcal{T} = \{t_1, t_2, \dots, t_m\}$ that includes $m$ noun phrases from WordNet~\cite{miller1995wordnet} and define a function $h(t_i)$ to obtain the embedding of each noun $t_i$ from CLIP~\cite{radford2021learning}, by constructing a sentence $s_i$ like ``$\texttt{A photo of a } \{t_i\}$'' and obtain their semantic embeddings as $\mathcal{V} = \{\mathbf{v}_1, \mathbf{v}_2, \dots, \mathbf{v}_m\}$ where $\mathbf{v}_i=h(s_i)$ from the text encoder of CLIP. Let $c$ be the number of categories; our goal is to group the images in $\mathcal{D}$ into $c$ clusters with the help of the CLIP model. Let $f(g(\mathcal{D});\phi):\mathcal{V}\rightarrow\mathbb{R}^c$ denotes the network with parameters $\phi$ that maps an image $x_i$ with embedding $\mathbf{u}_{i}$ into soft cluster assignment probability $\mathbf{q}_{i}$. $f$ is implemented by a multilayer perceptron (MLP). Notably, the image and text encoders in CLIP are kept frozen during the training process, i.e., the parameters in the functions $g(.)$ and $h(.)$ are fixed.

\section{The Proposed Method}

\begin{figure*}[t]
    \centering
    \includegraphics[width=0.8\linewidth]{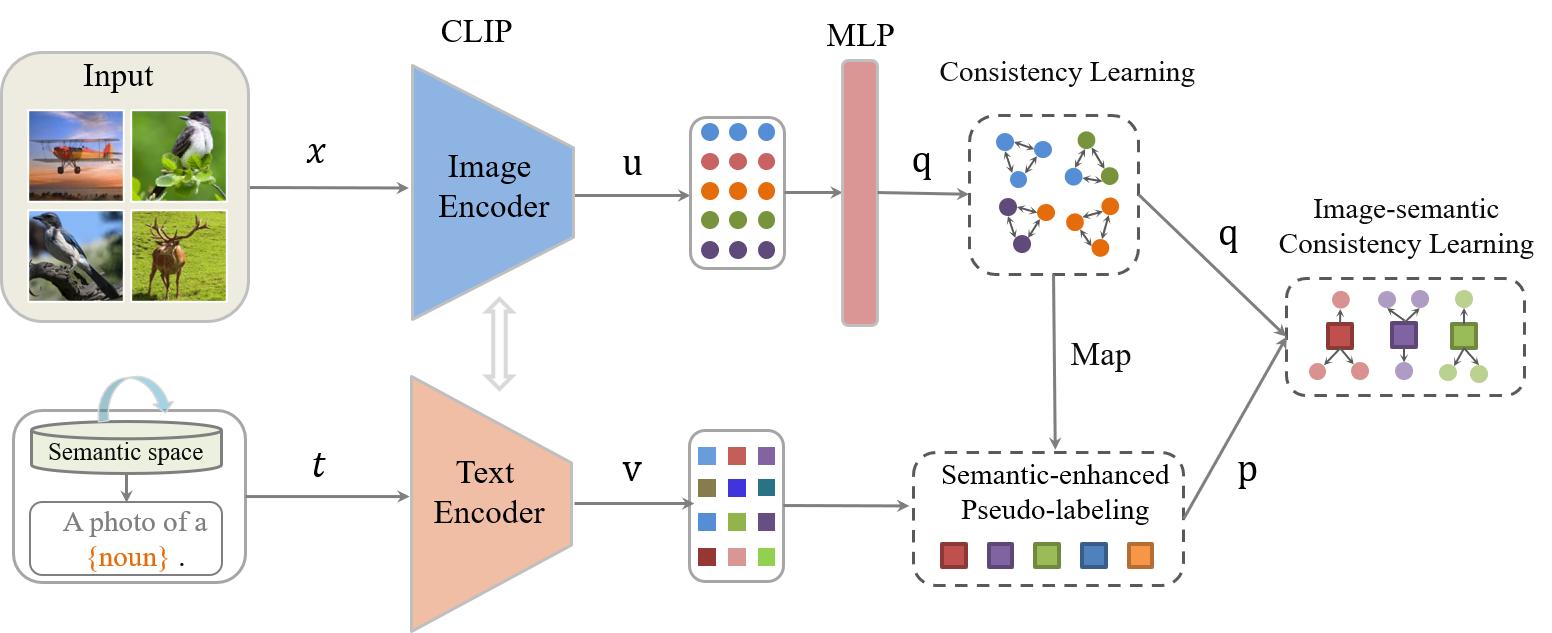}
    \caption{The framework of SIC, which consists of three parts: (1) Semantic space construction. (2) Semantic-Enhanced pseudo-labeling. (3) Joint consistency learning. Image features and semantic features are indicated by circle and square, respectively.}
    \label{fig:framework}
\end{figure*}

In this paper, we propose a novel image clustering method, which is shown in Figure~\ref{fig:framework}. The new method consists of three steps: 1) \textbf{Semantic Space Construction} selects meaningful texts to construct semantic space; 2) \textbf{Semantic-Enhanced Pseudo-labeling} generates pseudo-labels by taking both image and semantic spaces into consideration; and 3) \textbf{Joint Consistency Learning} performs image clustering with the consistency learning in both image and semantic spaces. In the following, we will give the details of the three steps.

\begin{comment}
\noindent \textbf{Multiple channels sharing assumption}(Assumption for the concept learning). \textit{Even though any individual channel contains other information, important factors are still perceivable since they tend to be shared among multiple channels.}
\end{comment}

\subsection{Semantic Space Construction}
In this step, we aim to construct a semantic space suitable for images by introducing related texts. In an image clustering task, we need to cluster images by their object category attributes, and the set of object category names is usually a subset of the commonly used nouns in the English language. For example, in CIFAR10, the class names are 10 commonly used English nouns (``airplane'', ``automobile'', ``bird'', etc.). Therefore, we take the entire list of nouns in the WordNet dataset~\cite{miller1995wordnet} to form a semantic dataset $\mathcal{W}$ which contains more than 82, 000 nouns. Since an image dataset usually covers only a small set of categories, we propose a two-step method to select most related nouns from $\mathcal{W}$. 

Some nouns contain a general meaning, i.e., ``object'', ``entity'', ``thing'', which will disturb the division of clusters. Intuitively, such nouns occur in most of the image-text pairs in training data and thus tend to locate near the text centers. Therefore, we compute a \textbf{uniqueness score} for each noun $\mathbf{w}\in\mathcal{W}$ as follows:
\begin{equation}
    \rho(\mathbf{w}) = 1-\frac{\mathbf{w}^{T}\mathbf{e}}{\|\mathbf{w}\|\|\mathbf{e}\|}
\end{equation}
where $\mathbf{e}=\frac{\sum_{\mathbf{w}\in\mathcal{W}}\mathbf{w}}{|\mathcal{W}|}$ is the text center.

We set a hyperparameter $\gamma_{u}$ to select that $\rho(\mathbf{w}) \geq \gamma_{u}$ as the most unique nouns by removing general worlds, resulting in a truncated noun subset $\mathcal{W}_{u} \in \mathcal{W}$ and $\rho(\mathbf{w}_{1})\geq\rho(\mathbf{w}_{2})$ holds for $\forall\mathbf{w}_1\in\mathcal{W}_{u}$ and  $\mathbf{w}_2\in\mathcal{W}-\mathcal{W}_{u}$.

Since the nouns in $\mathcal{W}_{u}$ may be irrelevant to the given images $\mathcal{D}$, we further filter $\mathcal{W}_{u}$ according to $\mathcal{D}$. Specifically, we first perform $k$-means clustering on $\mathcal{U}$ to obtain $c$ cluster centers and then select $\gamma_{r}$ nearest nouns for each cluster center to form the final semantic set $\mathcal{T}$ and their embeddings $\mathcal{V}=h(\mathcal{T})$. 

\subsection{Semantic-Enhanced Pseudo-labeling}

Thanks to the multi-modal pre-training models that bridge images and texts, we can connect images to semantics in an efficient way. Given the images $\mathcal{D}$ and their embeddings $\mathcal{U}$, this step aims to generate meaningful pseudo-labels according to the relationships between image embeddings $\mathcal{U}$ and semantic embeddings $\mathcal{V}$. To alleviate the above problem, we first generate $c$ representative semantic centers $\mathcal{H}$ and then generate the pseudo-labels $\mathcal{P}$ according to both $\mathcal{U}$ and $\mathcal{H}$.

We propose three strategies for generating the $c$ representative semantic centers $\mathcal{H}$: 

\noindent\textbf{1) Direct mapping.} This method directly maps each image $\mathbf{u}\in\mathcal{U}$ to its nearest semantic $\mathbf{v}\in\mathcal{V}$ where the dot product is the similarity function, and forms $\mathcal{S}$ as the nearest semantic set. Then we can perform $k$-means clustering on $\mathcal{S}$ to obtain $c$ cluster centers. 

\noindent\textbf{2) Center-based mapping.} Although the first method is very simple, it is cost-intensive and may result in ambiguous nearest semantics (see Figure~\ref{fig:motivation}) leading to meaningless representative semantic centers $\mathcal{H}$. Intuitively, if we cluster images according to the image features, the image cluster centers are more meaningful, and mapping the image cluster centers to semantics is more appealing. Given $\mathcal{Q}=f(g(\mathcal{D});\phi)$, we first select the top-$\xi_{c}$ images for each cluster by computing a binary matrix $\mathbf{Z}$, in which $z_{il}=1$ represents $x_i$ is selected as the top-$\xi_{c}$ samples for the $l$-th cluster, as follows:
\begin{equation}
    z_{il} =\left\{\begin{array}{l}
    1, \quad q_{il} \ge \kappa_l, \\
    0, \quad \text { otherwise }
    \end{array} 
    \right.
\end{equation}
where $\kappa_l$ is a dynamic threshold to cut the top branch samples according to cluster assignment probabilities as the epoch evolves, which is computed as:
\begin{equation}
    \kappa_l := \text{argtop-$\xi_{c}$}(\mathbf{\bar{q}}_l)
\end{equation}
where $\mathbf{\bar{q}}_l$ represents the $l$-th column of $\mathcal{Q}$.
Finally, we compute the image center $\mathcal{V}^{c}$ as follows:

\begin{equation}
    \mathbf{v}_{l}^{c}=\frac{1}{\left\|\mathbf{z}_{l}\right\|_{1}} \sum z_{i l} \mathbf{u}_{i}
\end{equation}
 After that, finding one semantic from $\mathcal{T}$ which is nearest to each image center in $\mathcal{V}^{c}$ results in the semantic centers $\mathcal{H}$.

\noindent\textbf{3) Adjusted center-based mapping.} Although the image cluster centers may map to more meaningful objects, the resulting semantic centers correspond to a set of nouns, which may limit the feasibility of the pseudo-labeling. In this method, we propose to recompute the semantic centers in $\mathcal{H}$ obtained by the second method. We first find $\xi_{a}$ nearest neighborhoods for each semantic $\mathbf{h}\in\mathcal{H}$ and then recompute the centers for each semantic as the final semantic centers $\mathcal{H}$.

\begin{comment}
Given a concept dataset $\mathcal{C}$, they can be formed as texts $\mathcal{T}$ with a fixed description ``\texttt{A photo of a}'', which are mapped into the text embedding features $\mathbf{e}_i$ through $g_{\mathcal{Z}}(t_i;\xi)$.  Obviously, suppose we have a image embedding feature $\mathbf{h}_i = f_{\mathcal{Z}}(v_i;\theta,\phi)$. let the nearest text embedding feature $\mathbf{e}_j$ to $\mathbf{h}_j$ as the concept of image. As shown in Figure~\ref{fig:image2concept}, images are mapped into the corresponding concepts in the text space through the CLIP model and each concept can accurately represent each image. However, such concepts are unable to generalize to the whole category and adaptive to the new coming images. Thus, this motivates us to learn a more robust concept for images.
\end{comment}

With the semantic centers $\mathcal{H}$, we propose an efficient way to generate the pseudo-labels. Given an image $x_{i}$, we first apply the dot product to measure the similarities between an image embedding $\mathbf{u}_i$ and $c$ semantic centers in $\mathcal{H}$ and then conduct a softmax operation following by an argmax operation to generate pseudo-labels $\mathcal{P}$ as follows:
\begin{equation}\label{def_pseudo_labels}
    \mathbf{p}_{i} = \operatorname{one-hot}\left(c,\operatorname{argmax}_{l}\frac{\exp \left(\mathbf{u}_{i}^T\mathbf{h}_{l} \right)}{\Sigma_{l^\prime}^{c} \exp \left(\mathbf{u}_{i}^T\mathbf{h}_{l^\prime} \right)}\right)
\text{}\end{equation}
where $\operatorname{one-hot}(c,l)$ will generate a $c$-bit one-hot vector with only one $1$ in the $l$-th position.

% \begin{equation}
%     z_{i l} =\left\{\begin{array}{l}
%     1, q_{i l} \ge \text{topk}_{i^\prime}\left(q_{i^\prime l}\right) \\
%     0, \quad \text { otherwise }
%     \end{array} 
%     \quad \text { s.t. topk }:= \left|C_{l}\right| * r\right.
% \end{equation}

\begin{comment}
Cross-modal semantic alignment:\\
\begin{equation*}
\begin{aligned}
u^{v} &=\left\{u_{j}^{v} \mid u_{j}^{v}=\frac{1}{\left\|z_{j}\right\|_{1}} \sum z_{i j} h_{i}, j \in[1, c], i \in[1, n]\right\} \\
z_{i j} &=\left\{\begin{array}{l}
1, q_{i j} \in \operatorname{topk}\left(C_{j}\right) \\
0, \quad \text { otherwise }
\end{array} \quad \text { s.t. topk }=\left|C_{l}\right| * r\right.\\
u^{t} &=\left\{u_{l}^{t} \mid u_{l}^{t} = \frac{1}{k} \sum_{i}^{k} e_{i}, e_{i} \in \mathcal{N}\left(u_{l}^{v}\right)\right\}
\end{aligned}    
\end{equation*}
\end{comment}

\subsection{Joint Consistency Learning}

Given an image $x_i$, we define its nearest neighborhood set as $\mathcal{N}_{k}(x_i)$, where $k$ is a predefined parameter for the nearest neighborhoods. To learn the model $f(g(\mathcal{D});\phi)$, we introduce the following assumptions for consistency learning:

\begin{assump}
    \textbf{Local smoothness assumption}(Assumption for the consistency learning). \textit{If two images $x_i$ and $x_j$ are located in a local neighborhood in the low-dimensional manifold, i.e, $x_j\in\mathcal{N}_{k}(x_i)$, then they have similar soft cluster assignments, i.e., $\mathbf{q}_i$ and $\mathbf{q}_j$ are similar.}
\end{assump}

\noindent\textbf{Image consistency learning:} according to the \textbf{local smoothness assumption}, we can learn the model $f(g(\mathcal{D});\phi)$ by enforcing the consistency between neighborhoods in the image space with the following loss:
\begin{equation}
\label{loss_cl}
\begin{split}
   \mathcal{L}_{I}(f(g(\mathcal{D});\phi)) =-\frac{1}{n} \sum_{i=1}^{n} \sum_{j=rn(\mathcal{N}_{k}(x_i))} \log \mathbf{q}_{i}^{T} \mathbf{q}_{j}
\end{split}
\end{equation}
where $\mathcal{N}_{k}(x_i)$ contains nearest neighbors of $x_i$ and $rn(\mathcal{N}_{k}(x_i))$ randomly selects a sample from $\mathcal{N}_{k}(x_i)$ for saving the computing cost.

\noindent\textbf{Image-semantic consistency learning:} With the generated pseudo-labels, we perform self-supervised learning of the model $f(g(\mathcal{D});\phi)$ with the following loss:
\begin{equation}\label{loss_iscl}
\mathcal{L}_{IS} = \frac{1}{n}\sum_{i=1}^{n} CE(\mathbf{p}_i,\mathbf{q}_i)    
\end{equation}
where $CE(.)$ is the cross entropy function.

Inspired by the $k$-meansNet~\cite{peng2018k}, we perform $k$-means clustering on $\mathcal{U}$ to obtain $c$ cluster centers as $\mathbf{R}\in\mathbb{R}^{c}$ to initialize the MLP parameters $\phi$ for reducing training time as follows:
\begin{align}
    \mathbf{W} & =  2\tau_{m}\mathbf{R} \\
    \mathbf{b} & = \{-\tau_{m} \|\mathbf{h}_l\|^{2}_{2}\}_{l=1}^{c}
\end{align}
where $\mathbf{W}$ and $\mathbf{b}$ are the weight and bias of MLP, and $\tau_{m}$ is the temperature parameter in the MLP model.

\noindent\textbf{Balance regularization:} We introduce the popular negative entropy loss for the balance clustering regularization, which can prevent the model from generating empty clusters:

\begin{equation}
\label{def_loss_b}
    \mathcal{L}_{B}(f(g(\mathcal{D});\phi))=-\sum_{l=1}^{c}\bar{q}_{l}\log \bar{q}_{l}
\end{equation}
where $\bar{q}_{l}=\frac{\sum_{i=1}^{n} q_{il}}{n}$ is the average cluster assignment.

% \subsection{Multi-task Co-learning}
\subsection{The Overall Objective}
The overall objective can be formulated as:
\begin{equation}\label{loss_overall}
\small
\begin{split}
    \min_{\phi}\mathcal{L}(f(g(\mathcal{D});\phi)) =& \min_{\phi} \mathcal{L}_{I}(f(g(\mathcal{D});\phi))+\beta\mathcal{L}_{IS}(f(g(\mathcal{D});\phi))\\ 
    &+\lambda\mathcal{L}_{B}(f(g(\mathcal{D});\phi))
\end{split}
\end{equation}
where $\beta$ and $\lambda$ are two trade-off parameters. 

\begin{algorithm}[t]
\caption{\textbf{Semantic-Enhanced Image Clustering} (SIC)}
\label{algorithm}
\KwIn{Images set $\mathcal{D}$, nouns set $\mathcal{W}$, neural networks $g(.)$, $h(.)$ and $f(.;\phi)$, training epoch $T$, cluster number $c$, hyperparameters $\gamma_u$ and $\gamma_r$, threshold $\kappa$, nearest neighborhoods number $k$, trade-off parameters $\lambda$ and $\beta$.}

\KwOut{Cluster assignments $\mathbf{Y}$.}
Update $\mathcal{U}=g(\mathcal{D})$ and $\mathcal{V}=h(\mathcal{T})$.

Filter $\mathcal{W}$ to obtain the semantic set $\mathcal{T}$ and embeddings $\mathcal{V}$ via \textbf{Semantic Space Construction}.

Initialize $\phi^{0}$ and $\mathcal{P}^{0}$.

\For{t = 0 to T}{
    % \For{$v_i \in \mathcal{V}$}{
    % $\mathcal{N}^{v}_{k} \leftarrow \mathcal{N}^{v}_{k}\cup \mathcal{N}^{v}_{k}(v_i)$, with $\mathcal{N}^{v}_{k}(v_i)$ contains the $k$ nearest neighbors of $v_i$ in the $f_{\mathcal{Z}}$ space.
    % }
% 	Sample a mini-batch $\{v_i\}_{i=1}^B$ from $\mathcal{V}$
    Update $\mathcal{Q}^{(t+1)}=f(g(\mathcal{D});\phi^{(t)})$.
    
    Generate $c$ representative semantic centers $\mathcal{H}$ from $\mathcal{U}$, $\mathcal{V}$ and $\mathcal{Q}^{(t+1)}$.
    
    Update pseudo-labels $\mathcal{P}^{t+1}$ via Eq. (\ref{def_pseudo_labels}).
    
    Update $\phi^{(t+1)}$ by optimizing Eq. (\ref{loss_overall}).
	
}
Output cluster assignments $\mathbf{Y}$ by $\mathbf{y}_{i}=\operatorname{one-hot}\left(\text{argmax}_{j}q^{(T+1)}_{ij}\right)$.
\end{algorithm}

\subsection{Theoretical Analysis}
In this part, we first analyze the convergence of our proposed method and then its expectation risk. Before analyzing, we first introduce the following assumptions 
\begin{assump}\label{assump:knn}
\textbf{Neighborhood Consistency Bound:}
    $\forall x_i \in \mathcal{X}$, $x_j \in \mathcal{N}_k(x_i)$, $\mathbf{q}_i^{T}\mathbf{q}_j \in[\mu_n,1].$ 
\end{assump}
\begin{assump}\label{assump:ce}
\textbf{Prediction Confidence Bound:}
    $\forall x_i \in \mathcal{X}$, $\|\mathbf{q}_i\|_{\infty} \leq \mu_p.$
\end{assump}
\begin{assump}\label{assump:num_neighbor}
\textbf{Neighborhood Imbalance Bound:}
    $\forall x_i \in \mathcal{X}$, $x_i$ is in at most $k'$ samples' (in $\mathcal{X}$) nearest neighborhoods.
\end{assump}

We first give the following theorem demonstrating that the optimization algorithm theoretically converges to the local optima (its proof is provided in the appendix due to space limitations.).

\begin{thm}
\label{thm:conv}
   Suppose that $f(.;\phi)$ and loss function $\mathcal{L}(f(g(\mathcal{D});\phi))$ are twice differential with bound gradients and Hessians. Besides, we assume that the loss function $\mathcal{L}(f(g(\mathcal{D});\phi))$ is
Lipschitz smooth with constant $L$. Suppose that the learning rate $\eta_{\phi}$ satisfies $\eta_{\phi}=\min \{\frac{1}{L}, \frac{C}{\sqrt{T}}\}$ for some $C>0$, such that $\frac{\sqrt{T}}{C} \geq L$. Then our proposed method can achieve $\min _{0 \leq t \leq T} \mathbb{E}\left[\left\|\nabla \mathcal{L}(g(\mathcal{D});\phi^{(t)})\right\|_{2}^{2}\right] \leq \epsilon$ in $\mathcal{O}\left(1 / \epsilon^{2}\right)$ steps, where $\epsilon$ is a very small positive real number.
\end{thm}

Next, we analyze the ability of our method to achieve cluster performance on unseen data. Let $\hat{\mathcal{L}}(f(g(\mathcal{D});\phi))$ be the empirical clustering risk of our method and its expectation can be donated as $\mathcal{L}(f(g(\mathcal{X});\phi))$. The family of $f$ is defined as $\mathcal{F}$. Recent works~\cite{liu2021refined,li2021sharper,tang2022deep} establish pioneering theoretical analysis for sharper generalization bound of clustering approaches. Inspired by these studies, we obtain the following theorem by analyzing the generalization bound of our proposed method (its proof is provided in the appendix due to space limitations.).

\begin{thm}
\label{thm:risk}
For any $0 < \delta < 1$, with at least probability $1-\delta$ for any $f\in\mathcal{F}$, the following inequality holds

\begin{equation*}
    \mathcal{L}(f(g(\mathcal{X});\phi))\leq \widehat{\mathcal{L}}(f(g(\mathcal{D});\phi)) + \frac{\tilde{c}_1}{\sqrt{n}} + \tilde{c}_2\sqrt{\frac{1}{2n}\log \frac{1}{\delta}}.
\end{equation*}
where $\tilde{c}_1=2\mu_{n}^{-1}+2C\beta+2c\lambda\log\mu_p^{-1}$ and $\tilde{c}_2= (2+2k')\log\mu_{n}^{-1} + C\beta +2c\lambda\log\mu_p^{-1}$. $C$ is a constant for the function $x\log x$. 
\end{thm}

Theorem~\ref{thm:risk} shows that our proposed method, with high probability $1-\delta$, is with a bounded expected clustering risk on the unseen data. To summarize, the proposed method is theoretically guaranteed to generalize clustering tasks. Note that $\mathcal{L}(f(g(\mathcal{X});\phi))$ is inversely proportional to $\mu_{n}$ and $\mu_{p}$ which reflect the neighborhood consistency and prediction confidence, indicating that improving the neighborhood consistency and prediction confidence reduces the expected risk. Meanwhile, $\mathcal{L}(f(g(\mathcal{X});\phi))$ is proportional to $k'$ which reflects the neighborhood overlapping, indicating that reducing the neighborhood imbalance (e.g., by setting a smaller number of neighbors $k$ or filtering neighborhoods to reduce neighborhood imbalance) also reduces the expected risk.

\section{Experiments and Analysis}
In this section, we conduct experiments on various public benchmark datasets to evaluate our proposed method.

\subsection{Experimental Setup}
\subsubsection{Datasets.}
We evaluated our method on five benchmark datasets, i.e. Cifar10~\cite{krizhevsky2009learning}, Cifar100-20~\cite{krizhevsky2009learning}, STL10~\cite{adam2011an}, ImageNet-Dogs~\cite{chang2017dog} and Tiny-ImageNet~\cite{ya2015tiny}. A brief description of these datasets is shown in Table~\ref{tab:data}. 

\begin{table}[htb]
  \begin{center}
    \normalsize
  \caption{Characteristics of five benchmark datasets.}
  \label{tab:data}
  \setlength{\tabcolsep}{0.5mm}{
  \begin{tabular}{c|cccc}\toprule
    Dataset & Image size & \#Training & \#Testing & \#Classes\\
  \hline
  \textbf{STL10} & $96\times96$ & $5,000$ & $8,000$ & $10$\\
  \textbf{Cifar10} & $32\times32$ & $50,000$ & $10,000$ & $10$\\
  \textbf{Cifar100-20} & $32\times32$ & $50,000$ & $10,000$ & $20$ \\
%   \textbf{ImageNet-10} & $224\times224$ & $13,000$ & $500$ & $10$ \\
  \textbf{ImageNet-Dogs} & $224\times224$ & $19,500$ & $750$ & $15$ \\
  \textbf{Tiny-ImageNet} & $64\times64$ & $100,000$ & $10,000$ & $200$ \\
  \bottomrule
  \end{tabular}}
  \end{center}
  \end{table}

\subsubsection{Evaluation metrics.}
We evaluate clustering results by three widely used metrics, including clustering Accuracy (ACC), Normalized Mutual Information (NMI)~\cite{mcdaid2011normalized} and Adjusted Rand Index (ARI)~\cite{hubert1985comparing}.

\subsubsection{Implementation details.}
For representation learning, we used the CLIP pre-training model, whose visual and text backbones are ViT-32~\cite{dosovitskiy2020image} and Transformer~\cite{vaswani2017attention}, separately. We obtained features from the image encoder of CLIP and then trained a cluster head. The cluster head is a fully connected layer with a size of $d\times c$, where $d$ and $c$ are the pre-training feature dimension and the number of clusters, respectively. During the training, the epoch numbers of all datasets were set to 100 with a batch size of 128. Before training, all datasets were augmented with the same method used in CLIP~\cite{radford2021learning}, i.e., a random square crop from resized images. The nearest neighbors were searched through Faiss Library ~\cite{jeff2021billion}. The best hyper-parameters used for five benchmark datasets are shown in Table ~\ref{tab:best_hyper}.

\begin{table}[t]
\begin{center}
\caption{The best hyper-parameters for each task. $\eta_{\phi}$: learning rate, $\gamma_{u}$: the number of most unique nouns, $\gamma_{r}$: the number of nearest nouns for each image center, $\xi_{c}$: the number of the top branch samples, $\xi_{a}$: the number of nouns nearest to the image center, $k$: the number of nearest neighbors in image consistency learning loss. $\lambda$ and $\beta$: trade-off parameters.}
\label{tab:best_hyper}
\centering
\resizebox{\linewidth}{!}{
\begin{tabular}{lcccccccc}
\toprule
\textbf{Dataset} & $\eta_{\phi}$ & $\gamma_{u}$ & $\gamma_{r}$ & $\xi_{c}$ & $\xi_{a}$& $k$ & $\lambda$ & $\beta$  \\
\bottomrule
STL10& 1e-4& 0.05 & 200 & 0.9$n/c$ & 20& 20  & 5 & 1 \\
Cifar10& 1e-4&  0.05   & 500 & 0.9$n/c$  &30 & 20 & 5 & 0.1  \\
 Cifar100-20& 1e-4 &  0.05  & 200 & 0.9$n/c$  &20 &20 & 5 & 1  \\
ImageNet-Dogs & 9e-3 &  0.05  & 1000 & 0.9$n/c$  &50 & 20 & 5 & 1   \\
Tiny-ImageNet & 1e-4 &  0.05   & 200 & 0.9$n/c$  &5 & 50 & 5 & 1  \\
\bottomrule
\end{tabular}
}
\end{center}
\end{table}

\subsection{Comparisons with State-of-the-art}

To evaluate the effectiveness of our proposed method, we compared it with 20 state-of-the-art clustering approaches on five datasets listed in Table~\ref{tab:data}. As shown in Table~\ref{tab:result}, different from most prior methods of training and evaluating the whole datasets on the top corner, we train and evaluate SCAN, NNM and SIC by using the train and val split respectively like SCAN~\cite{gansbeke2020scan}, which allows us to study the generalization properties of our method for novel unseen examples. The clustering results of six methods, i.e., SC~\cite{zelnik2005Self}, NMF~\cite{cai2009Locality}, AE~\cite{bengio2007greedy}, DAE~\cite{vincent2010stacked} and VAE~\cite{kingma2014autoencoding}, are obtained 
via $k$-means. 

Table~\ref{tab:result} shows the clustering results of our proposed method and the state-of-the-art methods on five benchmark datasets~\footnote{The clustering results (excluding those of our proposed method) are from the corresponding papers.}. It is clear that our proposed method outperforms all other methods on five datasets. Especially, our proposed method improves ACC, NMI and ARI by 17.2\%, 25.6\%, and 21.3\% on the STL10 dataset, 7.7\%, 10.7\%, and 10.7\% on the Cifar100-20 dataset and 19.2\%, 10.7\%, and 18.2\% on the Tiny-ImageNet dataset relative to the best results of all other methods. This means that our proposed method achieves a stable superior performance.

\begin{comment}
For clustering on the whole datasets, all training and testing data were combined together to train and evaluate models. For clustering on split datasets, the model trains the parameters through the training subset and then verifies the results through the verification subset. 
\end{comment}

\begin{table*}[t]
\centering
\caption{State-of-the-art comparison results on five benchmarks, including the averaged results of 5 different runs with standard deviation and the best model. The methods evaluation is divided into the whole dataset (top corner) and split datasets (bottom corner). We evaluated our proposed method on split datasets.The best results are shown in boldface.}
\setlength{\tabcolsep}{0.7mm}
\resizebox{\linewidth}{!}{
\begin{tabular}{@{}|l|ccc|ccc|ccc|ccc|ccc|ccc|@{}}
\toprule
Dataset &
  \multicolumn{3}{c|}{\textbf{STL10}} &
  \multicolumn{3}{c|}{\textbf{Cifar10}} & 
  \multicolumn{3}{c|}{\textbf{Cifar100-20}} & 
  \multicolumn{3}{c|}{\textbf{ImageNet-Dogs}} & 
  \multicolumn{3}{c|}{\textbf{Tiny-ImageNet}} \\ 
\midrule
Metrics &
  \multicolumn{1}{c}{ACC} &
  \multicolumn{1}{c}{NMI} &
  \multicolumn{1}{c|}{ARI} &
  \multicolumn{1}{c}{ACC} &
  \multicolumn{1}{c}{NMI} &
  \multicolumn{1}{c|}{ARI} &
  \multicolumn{1}{c}{ACC} &
  \multicolumn{1}{c}{NMI} &
  \multicolumn{1}{c|}{ARI} &
  \multicolumn{1}{c}{ACC} &
  \multicolumn{1}{c}{NMI} &
  \multicolumn{1}{c|}{ARI} &
  \multicolumn{1}{c}{ACC} &
  \multicolumn{1}{c}{NMI} &
  \multicolumn{1}{c|}{ARI} \\ 
\midrule

$k$-means~\cite{MacQueen-1967}      & 19.2 & 12.5 & 6.1  & 22.9 & 8.7   & 4.9  & 13.0  & 8.4  & 2.8  & 10.5 & 5.5 & 2.0 & 2.5 & 6.5 & 0.5\\
SC~\cite{zelnik2005Self}           & 15.9 & 9.8  & 4.8  & 24.7 & 10.3  & 8.5  & 13.6  & 9.0  & 2.2  & 11.1 & 3.8   & 1.3  & 2.2   & 6.3  & 0.4 \\
NMF~\cite{cai2009Locality}          & 18.0 & 9.6  & 4.6  & 19.0 & 8.1   & 3.4  & 11.8  & 7.9  & 2.6  & 11.8 & 4.4   & 1.6  & 2.9   & 7.2  & 0.5\\
               
% AC           & 33.2 & 23.9 & 14.0 & 22.8 & 10.5  & 6.5  & 13.8  & 9.8  & 3.4  & 13.9 & 3.7   & 2.1  & 2.7   & 6.9  & 0.5 \\
               
JULE~\cite{yang2016Joint}         & 27.7 & 18.2 & 16.4 & 27.2 & 19.2  & 13.8 & 13.7  & 10.3 & 3.3  & 13.8 & 5.4   & 2.8  & 3.3   & 10.2 & 0.6 \\
              
SAE~\cite{ng2011sparse}          & 32.0 & 25.2 & 16.1 & 29.7 & 24.7  & 15.6 & 15.7  & 10.9 & 4.4  & --   & --    & --   & --    & --   & --   \\
               
DAE~\cite{vincent2010stacked}          & 30.2 & 22.4 & 15.2 & 29.7 & 25.1  & 16.3 & 15.1  & 11.1 & 4.6  & 19.0 & 10.4  & 7.8  & 3.9   & 12.7 & 0.7 \\
             
% SWWAE        & 27.0 & 19.6 & 13.6 & 28.4 & 23.3  & 16.4 & 14.7  & 10.3 & 3.9 & --   & --    & --   & --    & --   & --\\
             
AE~\cite{bengio2007greedy}           & 30.3 & 25.0 & 16.1 & 31.4 & 23.4  & 16.9 & 16.5  & 10.0 & 4.7 & 18.5 & 10.4  & 7.3  & 4.1   & 13.1 & 0.7 \\
             
% DCGAN~\cite{radford2016unsupervised}        & 29.8 & 21.0 & 13.9 & 31.5 & 26.5  & 17.6 & 15.1  & 12.0 & 4.5 & 17.4 & 12.1  & 7.8  & 4.1   & 13.5 & 0.7  \\
              
% DeCNN        & 29.9 & 22.7 & 16.2 & 28.2 & 24.0  & 17.4 & 13.3  & 9.2  & 3.8  & 17.5 & 9.8   & 7.3  & 3.5   & 11.1 & 0.6  \\
              
VAE~\cite{kingma2014autoencoding}          & 28.2 & 20.0 & 14.6 & 29.1 & 24.5  & 16.7 & 15.2  & 10.8 & 4.0 & 17.9 & 10.7  & 7.9  & 3.6   & 11.3 & 0.6  \\
              
DEC~\cite{xie2016unsupervised}          & 35.9 & 27.6 & 18.6 & 30.1 & 25.7  & 16.1 & 18.5  & 13.6 & 5.0 & 19.5 & 12.2  & 7.9  & 3.7   & 11.5 & 0.7 \\
              
ADC~\cite{haeusser2018associative}          & 53.0 & --   & --  & 32.5 & --    & --   & 16.0  & --   & --  & --   & --    & --   & --    & --   & --   \\
             
DeepCluster~\cite{caron2018Deep}  & 33.4 & --   & -- & 37.4 & --    & --   & 18.9  & --   & -- & --   & --    & --   & --    & --   & --   \\
             
DAC~\cite{chang2017deep}          & 47.0 & 36.6 & 25.6 & 52.2 & 40.0  & 30.1 & 23.8  & 18.5 & 8.8 & 27.5 & 21.9  & 11.1 & 6.6   & 19.0 & 1.7 \\
             
DDC~\cite{chang2019deep}          & 48.9 & 37.1 & 26.7 & 52.4 & 42.4  & 32.9 & -- & --    & --  & --   & --    & --   & --     & --    & --   \\  
              
DCCM~\cite{wu2019deep}         & 48.2 & 37.6 & 26.2 & 62.3 & 49.6  & 40.8 & 32.7  & 28.5 & 17.3 & 38.3 & 32.1  & 18.2 & 10.8  & 22.4 & 3.8\\
              
IIC~\cite{ji2019invariant}          & 59.6 & 49.6 & 39.7 & 61.7 & 51.1  & 41.1 & 25.7  & 22.5 & 11.7 & --   & --    & --   & --    & --   & --\\
             
PICA~\cite{huang2020deep}         & 71.3 & 61.1 & 53.1 & 69.6 & 59.1  & 51.2 & 33.7  & 31.0 & 17.1 & 35.2 & 35.2  & 20.1 & 9.8   & 27.7 & 4.0\\

GCC~\cite{zhong2021graph}      & 78.8 & 68.4 & 63.1 & 85.6 & 76.4 & 72.8 & 47.2 & 47.2 & 30.5 & 52.6 & 49.0 & 36.2 & 13.8 & 34.7 & 7.5\\

CC~\cite{li2021contrastive}           & 85.0 & 76.4 & 72.6 & 79.0 & 70.5  & 63.7 & 42.9  & 43.1 & 26.6 & 42.9 & 44.5  & 27.4 & 14.0  & 34.0 & 7.1\\

\midrule
% Pretext + K-means & $65.8\pm5.1$ & $60.4\pm2.5$ & $50.6\pm4.1$ & $65.9\pm5.7$ & $59.8\pm2.0$ & $50.9\pm3.7$   & $39.5\pm1.9$ & $40.2\pm1.1$ & $23.9\pm1.1$ & $53.9\pm1.6$ & $56.6\pm0.9$ & $36.9\pm1.4$ & $34.8\pm0.2$ & $50.8\pm0.1$ &$66.7\pm0.1$\\
SCAN$^*$ (Avg{\tiny$\pm$Std})  & 75.5{\tiny$\pm$2.0} & 65.4{\tiny$\pm$1.2} & 59.0{\tiny$\pm$1.6} & 81.8{\tiny$\pm$0.3} & 71.2{\tiny$\pm$0.4} & 66.5{\tiny$\pm$0.4} & 42.2{\tiny$\pm$3.0} & 44.1{\tiny$\pm$1.0} & 26.7{\tiny$\pm$1.3} & 55.6{\tiny$\pm$1.5} & 58.7{\tiny$\pm$1.3} & 42.8{\tiny$\pm$1.3} & 41.1{\tiny$\pm$0.5} & 69.4{\tiny$\pm$0.3} & 32.7{\tiny$\pm$0.4}\\

SCAN$^\dagger$ (Avg{\tiny$\pm$Std}) & 76.7{\tiny$\pm$1.9} & 68.0{\tiny$\pm$1.2} & 61.6{\tiny$\pm$1.8}   & 87.6{\tiny$\pm$0.4} & 78.7{\tiny$\pm$0.5} & 75.8{\tiny$\pm$0.7} & 45.9{\tiny$\pm$2.7} & 46.8{\tiny$\pm$1.3} & 30.1{\tiny$\pm$2.1} & 59.2{\tiny$\pm$0.2} & 60.8{\tiny$\pm$0.4} & 45.3{\tiny$\pm$0.4} & \textbf{--}    & \textbf{--}   & \textbf{--}\\
SCAN$^\dagger$ (Best)~\cite{gansbeke2020scan} & 80.9 & 69.8 & 64.6  & 88.3 &79.7 & 77.2 & 50.7 & 48.6 & 33.3 & 59.3 & 61.2 & 45.7 & 42.0 & 69.8 & 33.2\\

NNM~\cite{dang2021nearest}          & 76.8{\tiny$\pm$1.2} & 66.3{\tiny$\pm$1.3} & 59.6{\tiny$\pm$1.5} & 83.7{\tiny$\pm$0.3} & 73.7{\tiny$\pm$0.5} & 69.4{\tiny$\pm$0.6} & 45.9{\tiny$\pm$0.2} & 48.0{\tiny$\pm$0.4} & 30.2{\tiny$\pm$0.4} & 58.6{\tiny$\pm$1.5} & 60.4{\tiny$\pm$0.5} & 44.9{\tiny$\pm$0.2} & 37.8{\tiny$\pm$0.1} & 66.3{\tiny$\pm$0.1} & 27.1{\tiny$\pm$0.1} \\

\textbf{SIC (direct) (Avg{\tiny$\pm$Std})} & 95.5{\tiny$\pm$0.1} & 92.7{\tiny$\pm$0.2} & 91.1{\tiny$\pm$0.2} & 78.3{\tiny$\pm$0.1} & 74.3{\tiny$\pm$0.1} &  66.9{\tiny$\pm$0.1} & 51.3{\tiny$\pm$0.1} & 53.9{\tiny$\pm$0.1} & 36.8{\tiny$\pm$0.1} & 59.0{\tiny$\pm$0.2} & 57.7{\tiny$\pm$1.8} & 41.1{\tiny$\pm$3.2} &  55.7{\tiny$\pm$0.8} & 77.4{\tiny$\pm$0.1}& 44.9{\tiny$\pm$0.6} \\

\textbf{SIC (center-based) (Avg{\tiny$\pm$Std})} & 96.7{\tiny$\pm$0.1} & 93.7{\tiny$\pm$0.1} & 93.2{\tiny$\pm$0.1} & 91.8{\tiny$\pm$0.1} & 83.4{\tiny$\pm$0.1} &  83.1{\tiny$\pm$0.1} & 54.0{\tiny$\pm$0.1} & 54.4{\tiny$\pm$0.4} & 38.6{\tiny$\pm$0.4} & 61.8{\tiny$\pm$1.1} & 63.9{\tiny$\pm$1.9} & 49.8{\tiny$\pm$1.4} & 61.0{\tiny$\pm$0.2} & 80.4{\tiny$\pm$0.1} & 51.2{\tiny$\pm$0.2} \\

\textbf{SIC (adjusted center-based) (Avg{\tiny$\pm$Std})} & 98.1{\tiny$\pm$0.1} & 95.3{\tiny$\pm$0.1} & 95.9{\tiny$\pm$0.1} & 92.6{\tiny$\pm$0.1} & 84.7{\tiny$\pm$0.1} &  84.4{\tiny$\pm$0.1} & 58.3{\tiny$\pm$0.1} & 59.3{\tiny$\pm$0.1} & 43.9{\tiny$\pm$0.1} & 69.7{\tiny$\pm$1.1} & 69.0{\tiny$\pm$1.6} & 55.8{\tiny$\pm$1.5} & 60.2{\tiny$\pm$0.3} & 79.4{\tiny$\pm$0.1} & 49.4{\tiny$\pm$0.2} \\

\textbf{SIC (Best)} & 
\textbf{98.1} & \textbf{95.4} & \textbf{95.9} & \textbf{92.7} & \textbf{84.8} & \textbf{84.6} & \textbf{58.4} & \textbf{59.3} & \textbf{44.0} & \textbf{71.3} & \textbf{71.8} & \textbf{58.6} & \textbf{61.2} & \textbf{80.5} & \textbf{51.4} \\
\bottomrule
\end{tabular}}
\label{tab:result}
\end{table*}

\subsection{Ablation Studies}
\subsubsection{Loss components effectiveness.}
We quantify the performance of loss components in our method through an ablation analysis, which consists of three losses: (a) the loss $\mathcal{L}_{I}$ for consistency between the image and its neighbor. (b) the loss $\mathcal{L}_{IS}$ for image-semantic consistency learning. (c) the loss $\mathcal{L}_{B}$ for the balance clustering regularization.
 Here we list the results on the Cifar10 in Table~\ref{tab:ablation_loss}. Both the losses $\mathcal{L}_{I}$ and $\mathcal{L}_{B}$ play a vital role in the overall performance improvement. Combine the loss $\mathcal{L}_{IS}$ to cluster together, the performance is improved by $8.2\%$, $6.8\%$ and $12.0\%$ in terms of  ACC, NMI and ARI, which indicates the effectiveness of our proposed image-semantic consistency learning.

\begin{comment}
Firstly, we conducted the loss $\mathcal{L}_{I}$ as a baseline. After that, we combined $\mathcal{L}_{I}$ and $\mathcal{L}_{IS}$ to cluster together, the performance is improved by $8.2\%$, $6.8\%$ and $12.0\%$ in terms of  ACC, NMI and ARI on Cifar10 dataset, which indicates the effectiveness of our proposed image-semantic consitency learning.

\end{comment} 
%     we first perform $\mathcal{L}_{I}$ and $\mathcal{L}_{IS}$ separately. $L_{cmcl}$ is able to obtain 8.2\%(ACC), 12.0\%(ARI), and 6.8\%(NMI) improvement, which indicates the pseudo-labels learning from text modal enhance the semantic discrimination power of model. 
%  Further, the $L_c$ achieve 21.4\%(ACC), 25.2\%(ARI), and 16.5\%(NMI) improvement, which shows consistency learning plays the most important role in the whole training process to force the sample and its neighbor in the same cluster.

\begin{table}[t]

  \begin{center}
    \normalsize
  \caption{Ablation studies of our method on Cifar10 dataset.}
  \label{tab:ablation_loss}
  \setlength{\tabcolsep}{4mm}
  \resizebox{\linewidth}{!}{
  \begin{tabular}{c|ccc}\toprule
   Setup & ACC & NMI & ARI\\
  \hline
  \textbf{w/o} $ \mathcal{L}_{IS}$ & $84.4\pm0.5$ & $77.9\pm0.3$ & $72.4\pm0.5$ \\
  \textbf{w/o}$ \mathcal{L}_{I}$ & $71.2\pm0.1$ & $68.2\pm0.2$  & $59.2\pm0.3$\\
  \textbf{w/o}$ \mathcal{L}_{B}$ &$70.3\pm7.2$ & $74.6\pm2.5$ & $58.8\pm6.0$  \\
%   \textbf{w/o} $\mathcal{L}_{b}$ & $91.1\pm0.0$ &  $82.9\pm0.0$ & $81.4\pm0.0$\\
  $\textbf{SIC}$ & $\textbf{92.6}\pm\textbf{0.1}$ &$\textbf{84.7}\pm\textbf{0.1}$ & $\textbf{84.4}\pm\textbf{0.1}$  \\  
  \bottomrule
  \end{tabular}}
  \end{center}
\end{table}

\subsubsection{Comparison on three semantic mapping methods.}
We also conduct experiments to compare the three methods for mapping images to semantic centers, i.e., \textbf{direct mapping}, \textbf{center-based mapping} and \textbf{adjusted center-based mapping}.
As shown in Table~\ref{tab:result}, SIC with adjusted center-based mapping achieves the best results, and SIC with direct mapping achieves the worst results.

\begin{figure}[!htb]
\centering
    \subfloat[\footnotesize ImageNet-Dogs.]{
           \centering 
           \label{fig:imagenet_dog_pl_acc}  
           \includegraphics[width = .5\linewidth]{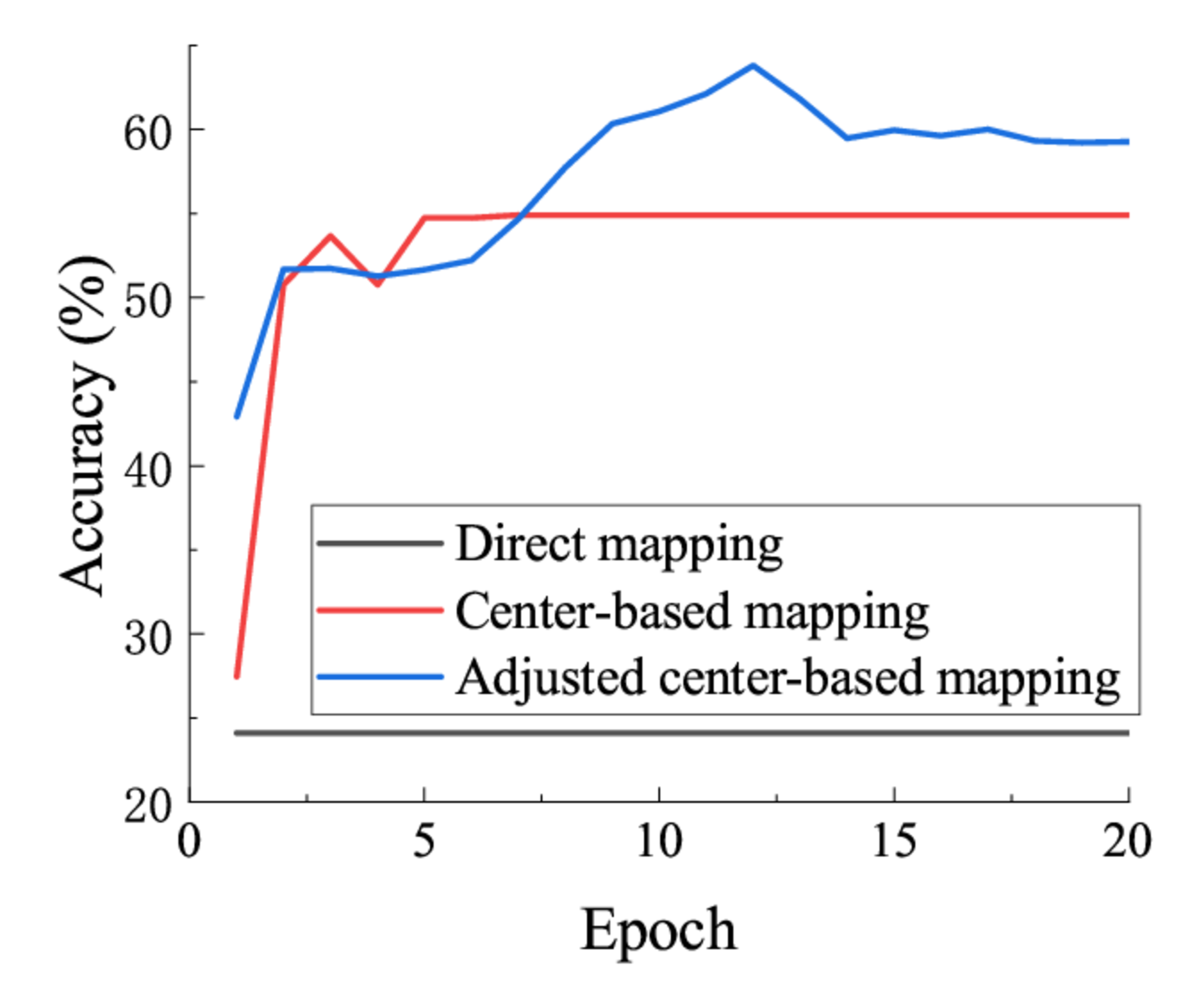}}\hspace{-5mm}
    \subfloat[\footnotesize STL10.]{
            \centering 
            \label{fig:stl10_pl_acc}
            \includegraphics[width = .5\linewidth]{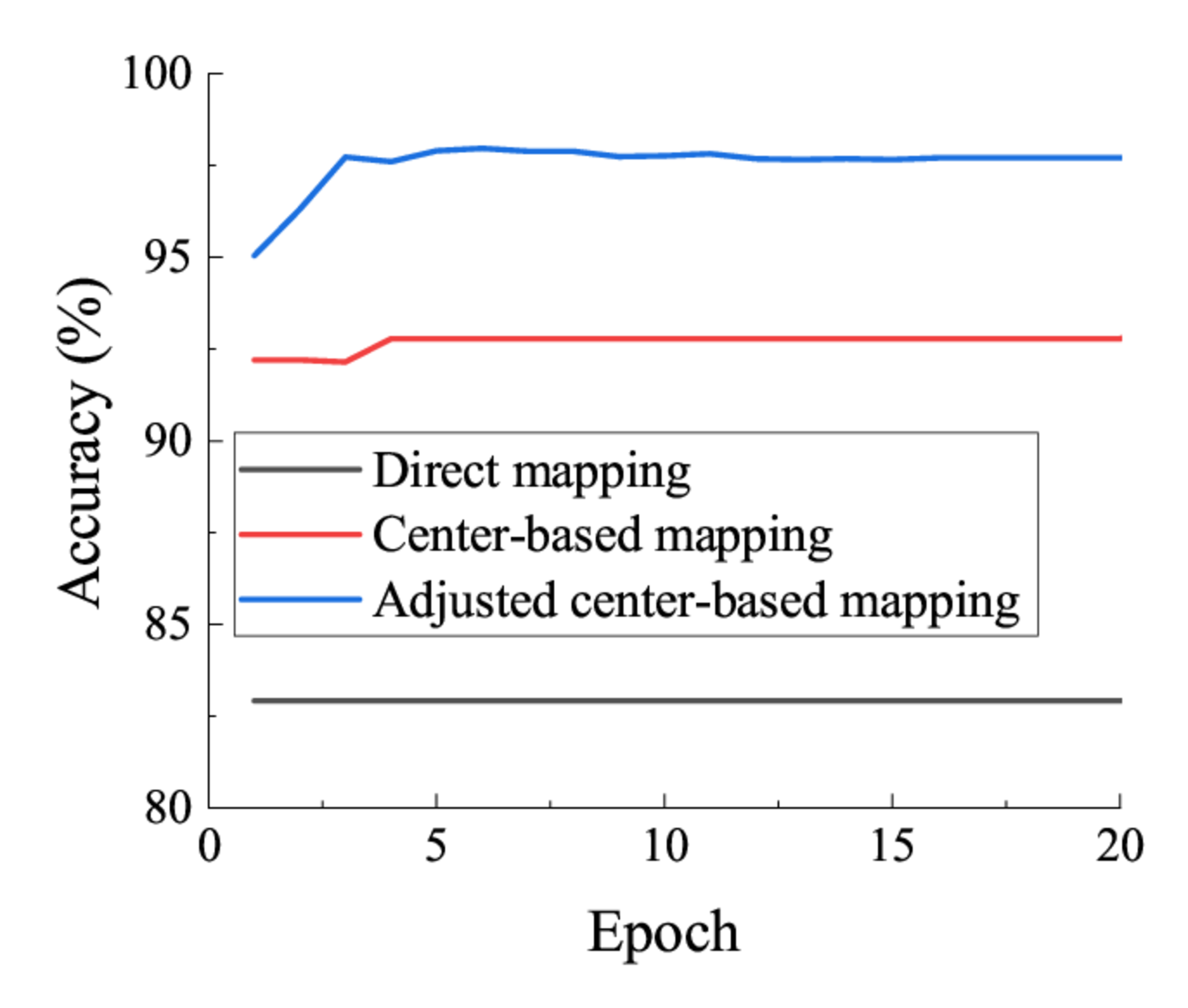}}
            
\caption{The accuracy of pseudo-labels as epoch evolves on ImageNet-Dogs and STL10 datasets.}
\label{fig:pl_acc}
\end{figure}

We also investigate the quality of pseudo-labels generated by three methods, i.e., \textbf{direct mapping}, \textbf{center-based mapping} and \textbf{adjusted center-based mapping}. As shown in Figure~\ref{fig:pl_acc}, we can observe that SIC with adjusted center-based mapping performs best while SIC with direct mapping performs worst.

The above results verify that direct mapping each image to its nearest semantic and performing $k$-means to obtain centers are not good and result in low-quality pseudo-labels. Moreover, applying the adjusted centers improves the semantic centers and pseudo-labels.

\subsubsection{Compared to relative results of CLIP.}
To display the clustering power of our model, we compare SIC with ``CLIP+zero-shot'' and ``CLIP+$k$-means'' on the STL10, Cifar10, and ImageNet-Dogs datasets. ``CLIP+zero-shot'' uses the given class names in each dataset to directly classify images with CLIP, and  ``CLIP+$k$-means'' performs $k$-means clustering on image embeddings obtained by the image encoder in CLIP. In Table~\ref{tab:cmp_clip}, it is clear that SIC outperforms the other two methods, indicating that our method can better utilize CLIP to uncover image clusters without class names.
\begin{table}[t]
\begin{center}
\centering
\caption{
Ablation studies of our method compared to ``CLIP+zero-shot'' and ``CLIP+$k$-means''.  
}
\label{tab:cmp_clip}
\setlength{\tabcolsep}{4mm}
\resizebox{\linewidth}{!}{
\begin{tabular}{c|ccc}\toprule
Methods (ACC)& 
STL10 & Cifar10& ImageNet-Dogs \\
\hline
CLIP+zero-shot & 95.7 & 80.0 & 34.1\\
CLIP+$k$-means & 94.6{$\pm$0.1} & 75.3{$\pm$0.1} & 39.8{$\pm$3.9} \\ 
\textbf{SIC}  & \textbf{98.1{$\pm$0.1}} & \textbf{92.6{$\pm$0.1}} & \textbf{69.7{$\pm$1.1}} \\
\bottomrule
\end{tabular}
}
\end{center}

\end{table}

\subsubsection{Visualization of learned image features.}

 \begin{comment}
   \begin{figure}[!htb]
\centering
    \subfloat[\footnotesize Image features]{
           \centering 
           \label{fig:image_tsne}  
           \includegraphics[width = .5\linewidth]{figures/image_tsne.png}}
    \subfloat[\footnotesize Nearest text features]{
            \centering 
            \label{fig:concept_tsne}
            \includegraphics[width = .5\linewidth]{figures/concept_tsne.png}}
\caption{$t$-SNE visualization of image features and nearest text features of STL10 by CLIP.}
\label{fig:image_to_semantic}
\end{figure}

Figure~\ref{fig:image_to_semantic} visualizes the image features obtained by CLIP, from which we can observe that the text features show ambiguous cluster structures.
\end{comment}

\begin{figure*}[!htb]
\centering
    \subfloat[{\small CLIP}]{
           \centering 
           \label{fig:tsne_clip}  
           \includegraphics[width = .25\linewidth]{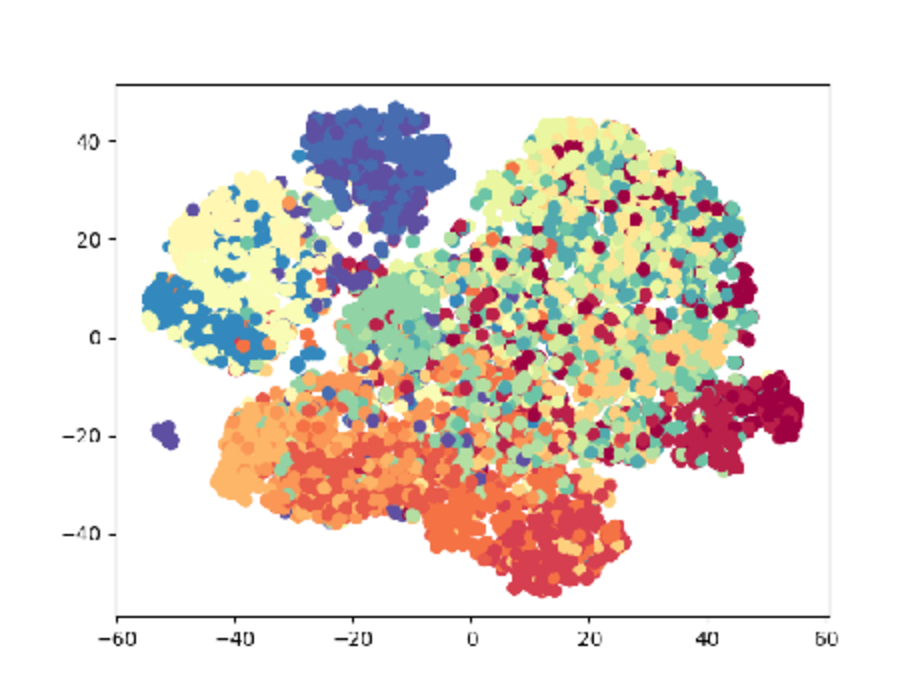}}
    \hspace{0.05\linewidth}
    \subfloat[{\small Image consistency learning}]{
            \centering 
            \label{fig:tsne_i}
            \includegraphics[width = .25\linewidth]{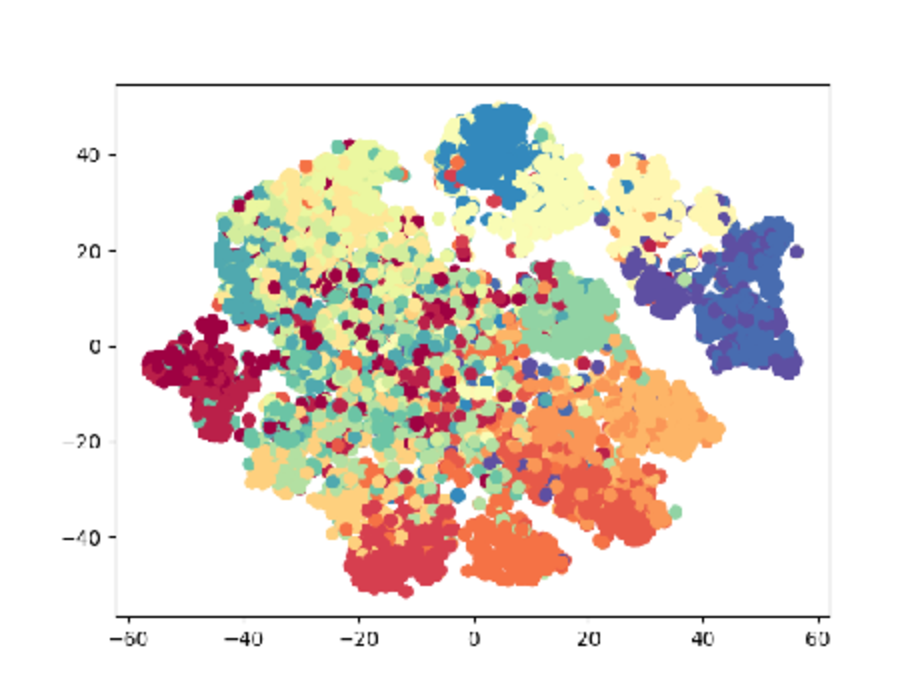}}
    \hspace{0.05\linewidth}
    \subfloat[{\small SIC}]{
            \centering 
            \label{fig:tsne_sic}
            \includegraphics[width = .25\linewidth]{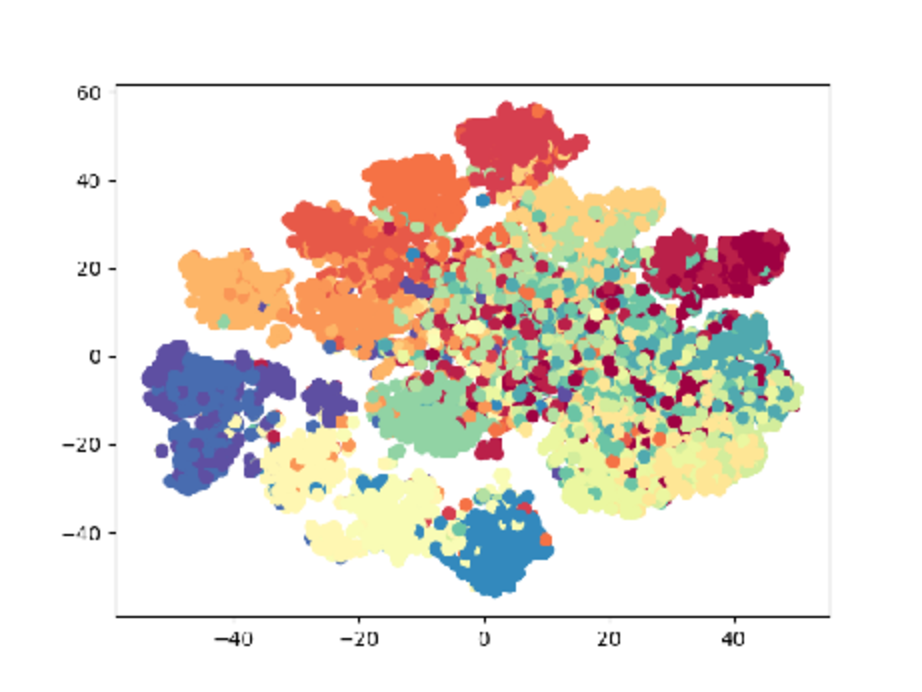}}
            
\caption{$t$-SNE visualization of learned image features from CLIP, image consistency learning, and SIC on the Cifar100-20 dataset.}
\label{fig:tsne_cifar20}
\end{figure*}

Figure~\ref{fig:tsne_cifar20} visualizes the image features obtained by CLIP, image consistency learning (before softmax), and SIC (before softmax) by $t$-SNE on the Cifar100-20 dataset. We can observe ambiguous cluster structures from the image features obtained by CLIP. Although image consistency learning improves image embeddings, we also observe ambiguous cluster structures. However, with our proposed method, we can observe the clearest structures.

\subsection{Sensitivity Analysis}

\begin{figure}[!htb]
\centering
    \subfloat[\footnotesize ($\lambda$,$\beta$) on Cifar10.]{
           \centering 
           \label{fig:cifar10_lossweight}  
           \includegraphics[width = .46\linewidth]{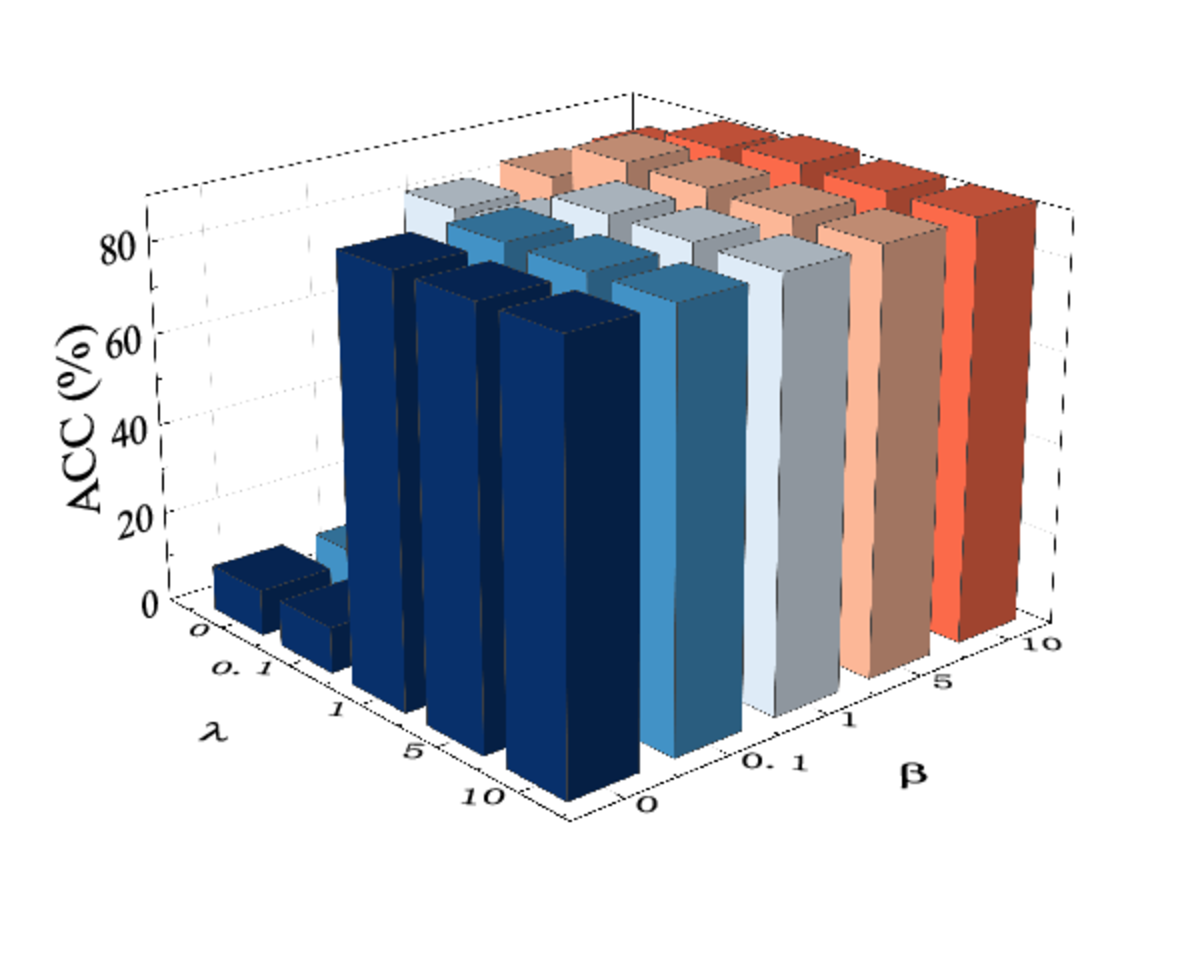}}\hspace{-3mm}
    \subfloat[\footnotesize ($\lambda$,$\beta$) on ImageNet-Dogs.]{
            \centering 
            \label{fig:imagenetdog_lossweight}
            \includegraphics[width = .46\linewidth]{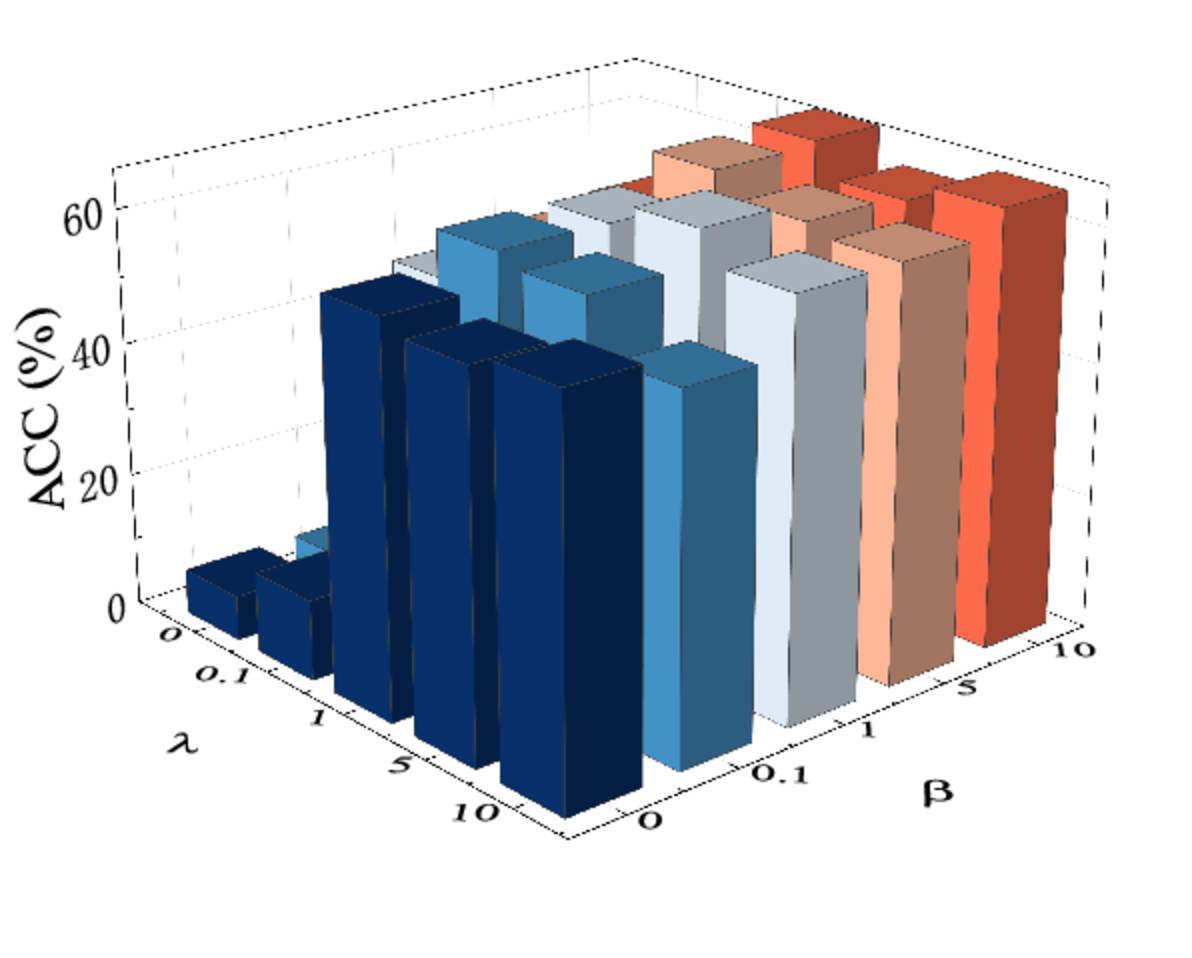}}
            
\caption{Sensitivity analysis of $\lambda$ and $\beta$.}
\label{fig:sens_trade_off}
\end{figure}

\noindent \textbf{Sensitivity on trade-off parameters $\lambda$ and $\beta$.} We study the influence of trade-off parameters $\lambda$ and $\beta$, where $\beta$ helps to separate the visually similar but semantically different images and  $\lambda$ helps prevent the model into a trivial solution. We set $\lambda,\beta \in [0, 0.1, 1, 5, 10]$ to show the sensitivity results in Figure~\ref{fig:sens_trade_off}. In general, decreasing $\lambda$ causes performance degradation, and increasing $\beta$ improves performance.

\begin{figure}[!htb]
\centering
    \subfloat[\footnotesize $\gamma_u$.]{
            \centering 
            \label{fig:cifar10_gamma_u}
            \includegraphics[width = .5\linewidth]{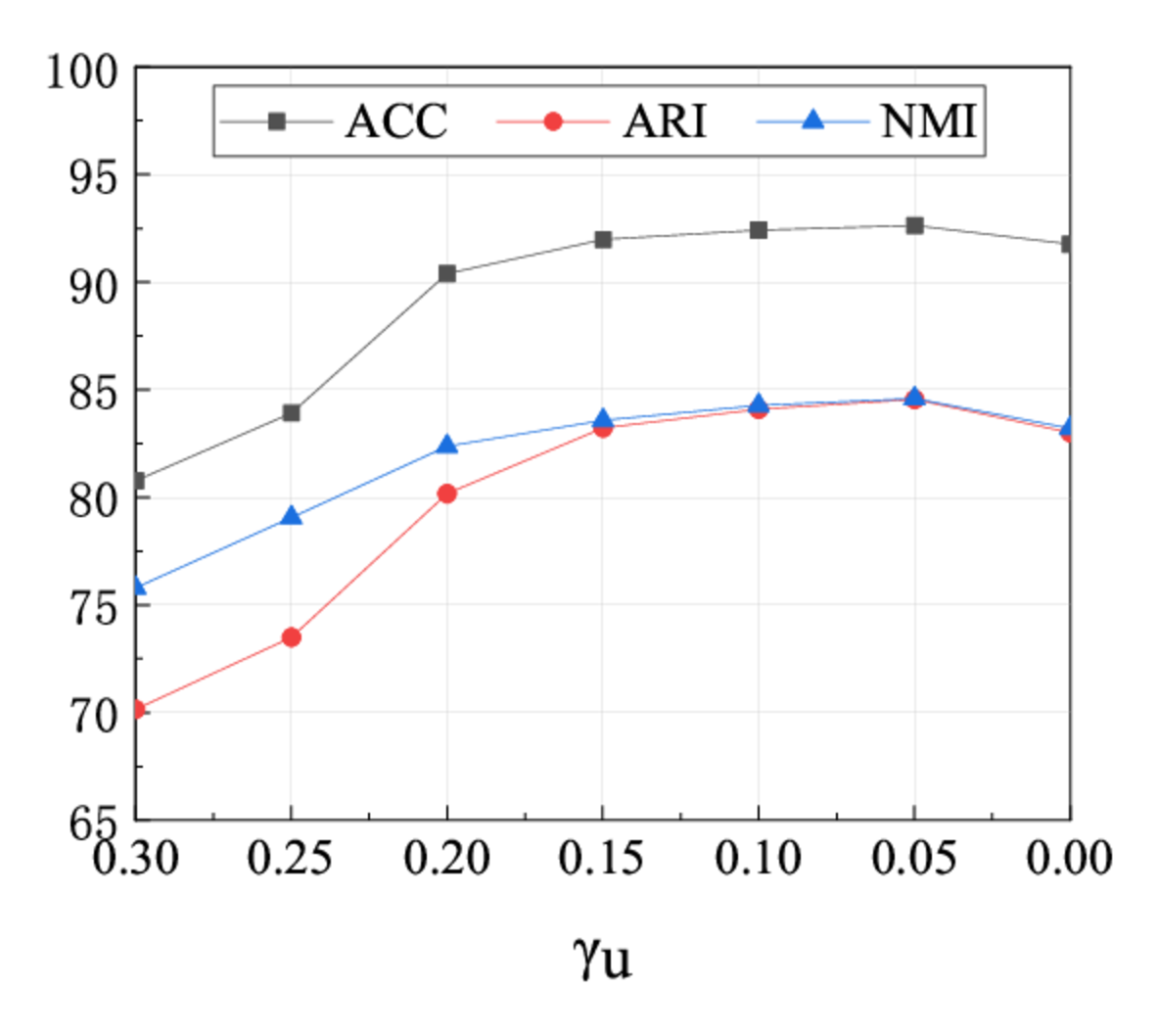}}
    \subfloat[\footnotesize $\gamma_r$.]{
           \centering 
           \label{fig:cifar10_gamma_r}  
           \includegraphics[width = .5\linewidth]{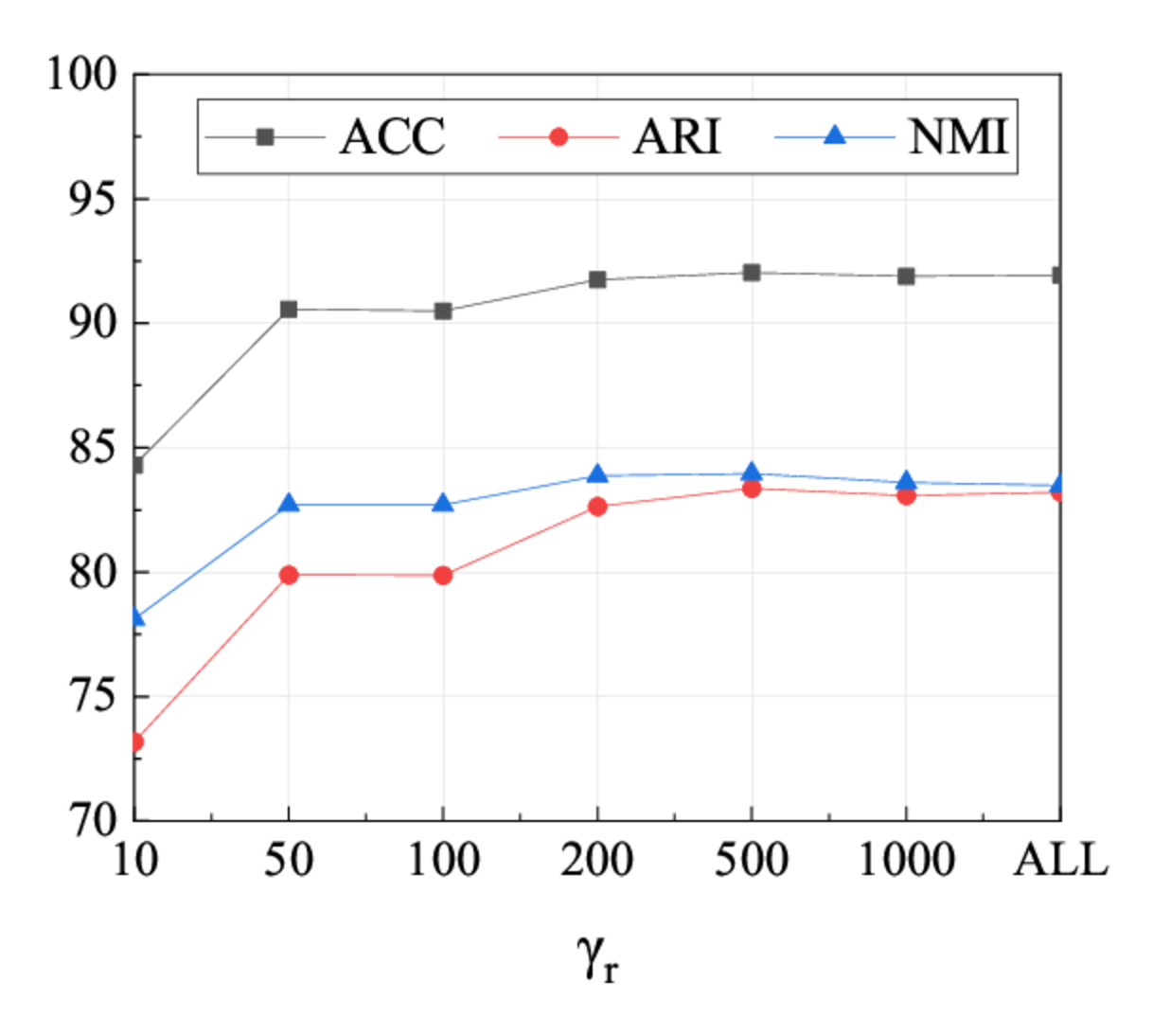}}\hspace{-5mm}    
\caption{Sensitivity analysis of $\gamma_u$ and $\gamma_r$ on Cifar10.}
\label{fig:sens_gamma}
\end{figure}

\noindent \textbf{Sensitivity on hyperparameters $\gamma_u$ and $\gamma_r$.} In our method, $\gamma_u$ and $\gamma_r$ are used to select proper nouns from WordNet by removing the general words. As shown in Figure~\ref{fig:sens_gamma}, we can observe that decreasing $\gamma_u$ and increasing $\gamma_r$ improves the performance first, then does not improve the performance too much, indicating that removing general worlds can reduce the computing cost without performance degeneration too much. We also observe that decreasing $\mu_u$ from $0.05$ to $0$ causes performance degradation, indicating that removing general words is necessary.

%  In Figure~\ref{fig:cifar10_nc}, shows the performance of the different value. The result indicates that the clustering performance improved by noun counts when $\gamma_r \leq 200$, and remains stable when $\gamma_r >200$. Figure~\ref{fig:imagenet_dog_nc} indicates $\gamma_r$ improves the clustering result when $\gamma_r \leq 200$, and decreases the result when $\gamma_r > 200$.Therefore, $\gamma_r = 200$ is a good choice for imagenet-dogs dataset. $\gamma_r$ for other datasets do like this.

\begin{figure}[!htb]
\centering
    \subfloat[\footnotesize $\xi_{a}$ on Cifar10.]{
           \centering 
           \label{fig:cifar10_ck}  
           \includegraphics[width = .5\linewidth]{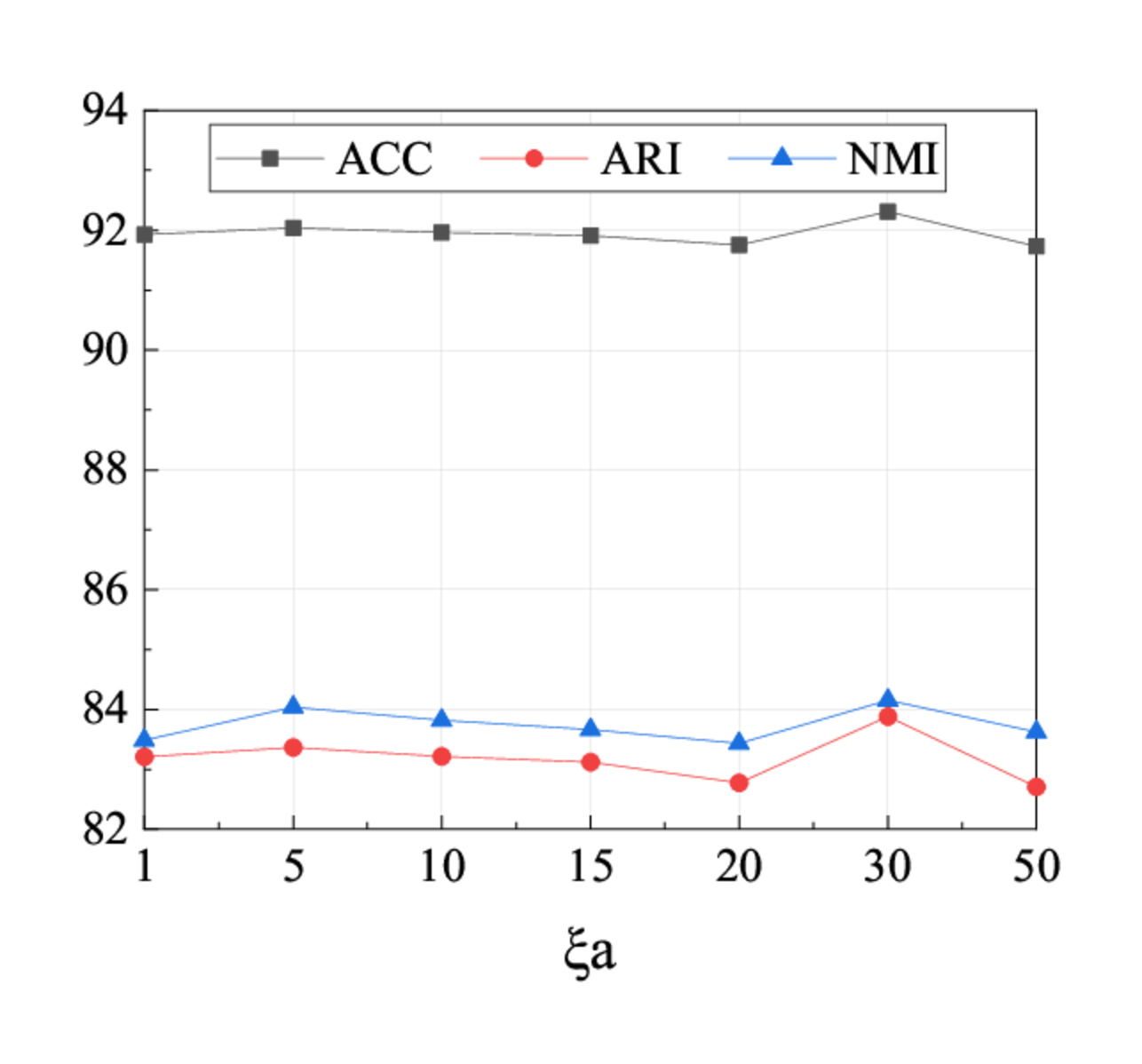}}\hspace{-5mm}
    \subfloat[\footnotesize $\xi_{a}$ on ImageNet-Dogs.]{
            \centering 
            \label{fig:imagenet_dog_ck}
            \includegraphics[width = .5\linewidth]{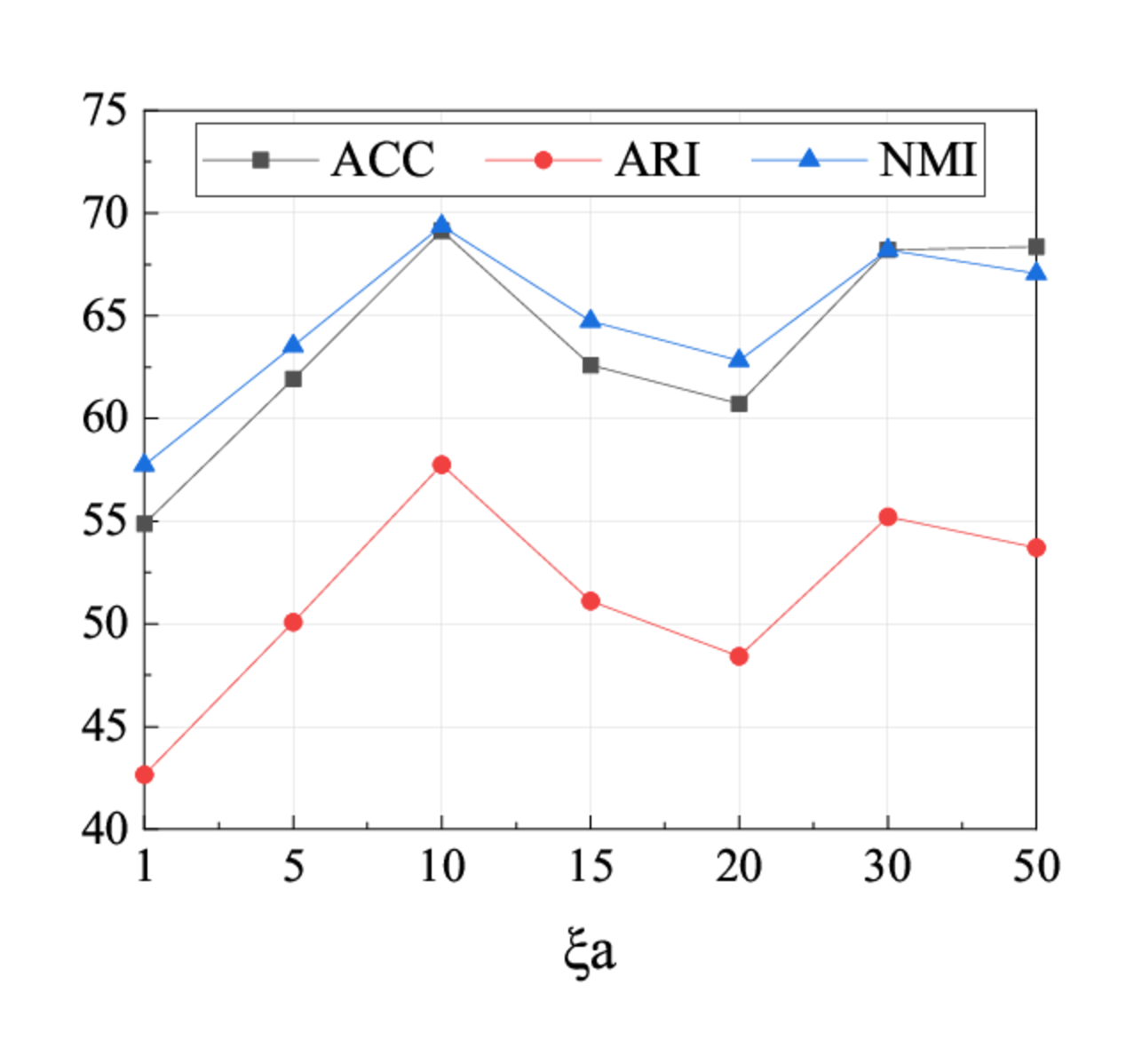}}
            
\caption{Sensitivity analysis of $\xi_{a}$.}
\label{fig:sens_xi_a}
\end{figure}

\begin{figure}[!htb]
\centering
    \subfloat[\footnotesize $\xi_c$ on Cifar10.]{
           \centering 
           \label{fig:cifar10_ratio}  
           \includegraphics[width = .5\linewidth]{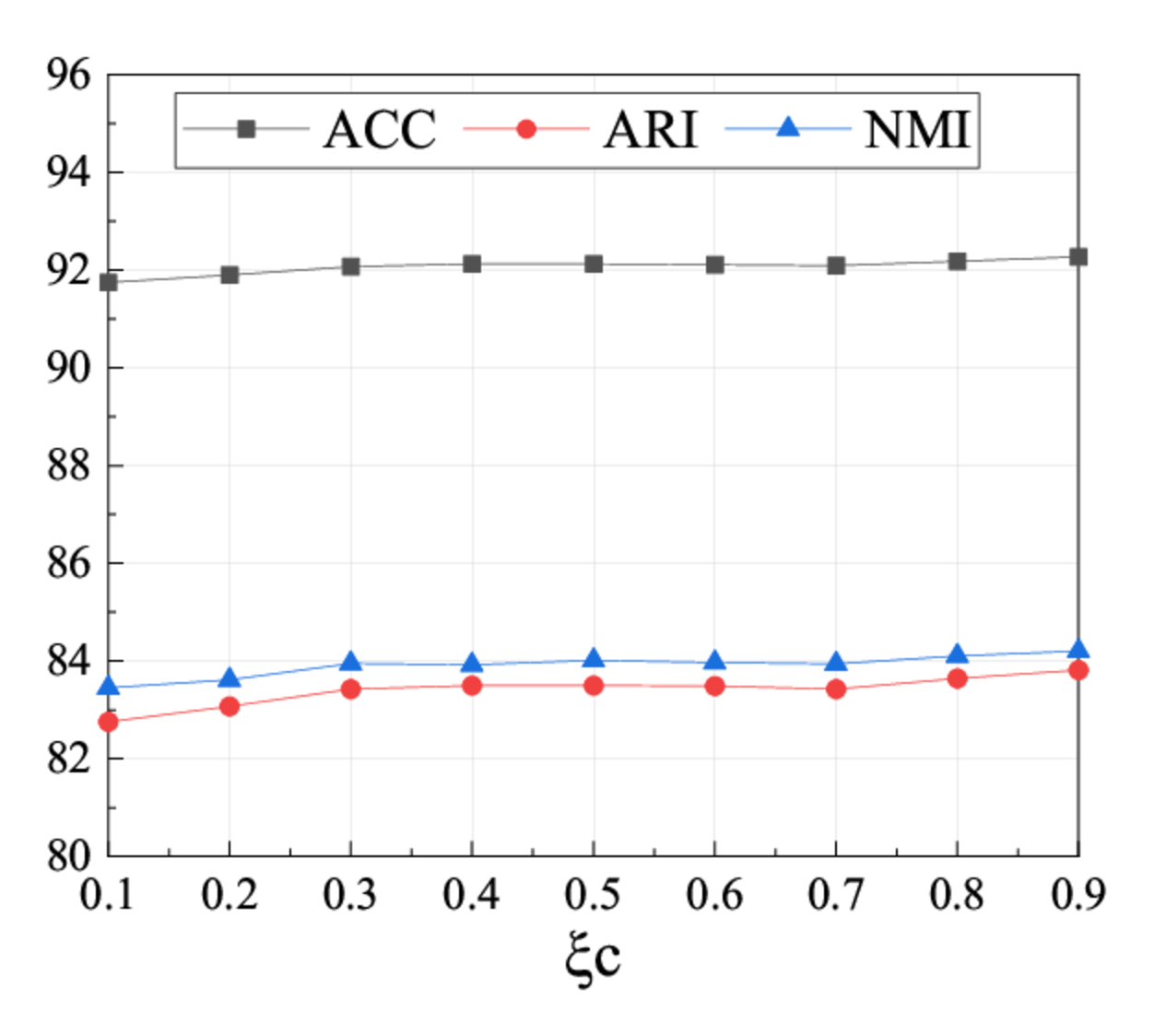}}\hspace{-5mm}
    \subfloat[\footnotesize $\xi_c$ on Imagenet-dogs.]{
            \centering 
            \label{fig:imagenet_dog_ratio}
            \includegraphics[width = .5\linewidth]{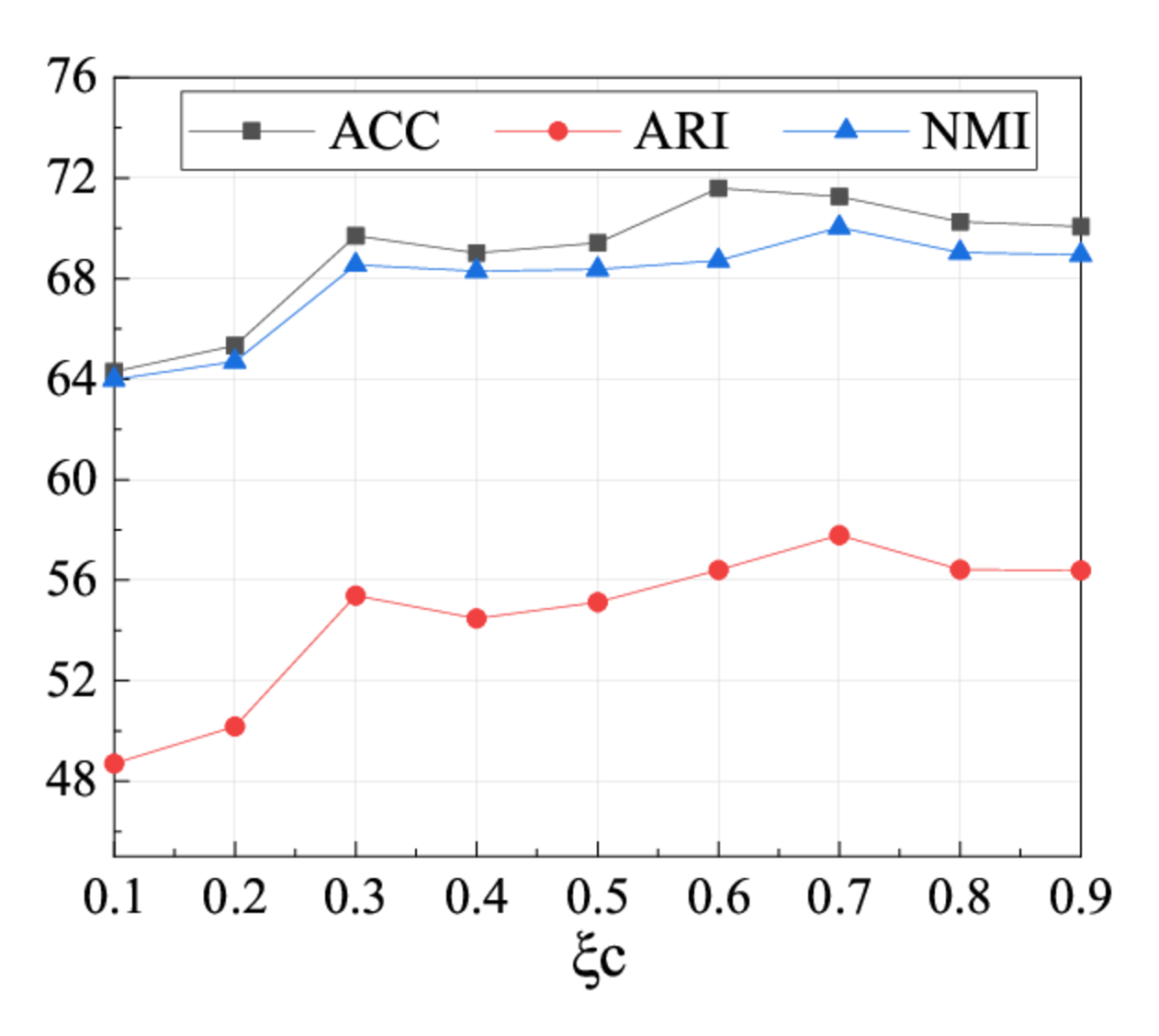}}
            
\caption{Sensitivity analysis of $\xi_c$.}
\label{fig:sens_xi_c}
\end{figure}

\noindent \textbf{Sensitivity on hyperparameters $\xi_{a}$ and $\xi_{c}$.} $\xi_{a}$ is used to adjust the semantic centers and $\xi_c$ is used to select the top branch samples. Figures~\ref{fig:sens_xi_a} and \ref{fig:sens_xi_c} show the sensitivity results on both $\xi_{a}$ and $\xi_{c}$, respectively. We can observe different sensitives of $\xi_{a}$ and $\xi_{c}$ on different datasets. For example, $\xi_{a}$ and $\xi_{c}$ do not affect the performance too much on the Cifar10 dataset, but affect too much on the ImageNet-Dogs dataset.

\begin{comment}
we trained with different settings of $c_k$, where $c_k \in [1, 5, 10, 15, 20, 30, 50]$.  Figure~\ref{fig:imagenet_dog_ck} shows the increasing $c_k$ improves the clustering performance when $c_k \leq 10$. As the result shown in Figure~\ref{fig:sens_xi_a}, different datasets have different sensitivity to $c_k$, where Cifar10 performs stable but Imagenet-dog perform fluctuated. We find $c_k = 30$ in Cifar10 and $c_k = 10$ in Imagenet-dogs achieves the best performance.  
\end{comment}

\section{Conclusion}
This paper proposes a novel image clustering \textbf{SIC} which utilizes the visual-language pre-training model CLIP to compensate the semantic information for better image clustering. We propose efficient methods to map images to a proper semantic space and cluster images from both image and semantic spaces. Theoretical results show that SIC can converge and reveal that the expected risk of SIC is affected by the models' performance in terms of neighborhood consistency and prediction confidence. The imbalance of the constructed neighborhoods also affects the expected risk of SIC. Experimental results show that our method outperforms 20 state-of-the-art and zero-shot learning with CLIP, enabling its wide potential applications. However, the pseudo-labels generated in our method may be suboptimal, so we will study new methods to generate better pseudo-labels. It is deserved to extend our theoretical results to self-supervised learning.

\newpage

\section*{Acknowledgments}
This work is jointly supported by Major Project of the New Generation of Artificial Intelligence (No. 2018AAA0102900); in part by NSFC under Grant no. 92270122; and in part by the Shenzhen Research Foundation for Basic Research, China, under Grant JCYJ20210324093000002.

% \newpage

%\appendix
%\section{References}
%\label{sec:reference_examples}

\nobibliography*

% Use \bibliography{yourbibfile} instead or the References section will not appear in your paper
\bibliography{clustering,deepclustering,representation,image,foundation,chen,crossmodel}

\onecolumn
\section{Appendix}
\subsection{Proofs}
In this appendix, we provide the detailed proofs of the theoretical results. 

\subsection{Proof of Theorem 1}
We first reformulate the update rule as \\
$$ \phi^{(t+1)}=\phi^{(t)}-\eta_{\phi}(\nabla \mathcal{L}(\phi^{(t)})+\zeta^{(t)}),$$
\noindent where $\zeta^{(t)}=\left.\nabla \mathcal{L}(\phi^{(t)})\right|_{\mathcal{B}}-\nabla \mathcal{L}(\phi^{(t)})$ and $\mathcal{B}$ is a mini-batch data sampled i.i.d from $\mathcal{X}$. This indicates that $\mathbb{E}[\zeta^{(t)}]=0$ holds. Let $\|\zeta^{(t)}\|^2 \leq \sigma^2$. First, since $\mathcal{L}$ is Lipschitz smooth with constant $L$, we have \\
$$
\begin{aligned}
& \mathcal{L}(\phi^{(t+1)})-\mathcal{L}(\phi^{(t)}) \\
\leq & \langle\nabla \mathcal{L}(\phi^{(t)}), \phi^{(t+1)}-\phi^{(t)}\rangle+\frac{L}{2}\|\phi^{(t+1)}-\phi^{(t)}\|_{2}^{2}\\
=&\langle\nabla \mathcal{L}(\phi^{(t)}),-\eta_{\phi}[\nabla\mathcal{L}(\phi^{(t)})+\zeta^{(t)}]\rangle+\frac{L \eta_{\phi}^{2}}{2}\|\nabla \mathcal{L}(\phi^{(t)})+\zeta^{(t)}\|_{2}^{2} \\
=&-(\eta_{\phi}-\frac{L_{\phi}^{2}}{2})\|\nabla \mathcal{L}(\phi^{(t)})\|_{2}^{2}+\frac{L \eta_{\phi}^{2}}{2}\|\zeta^{(t)}\|_{2}^{2}-(\eta_{\phi}-\operatorname{L\eta }_{\phi}^{2})\langle\nabla \mathcal{L}(\phi^{(t)}), \zeta^{(t)}\rangle .
\end{aligned}
$$

\noindent Therefore, \\
$$
\begin{aligned}
&(\eta_{\phi}-\frac{L\eta_{\phi}^{2}}{2})\|\nabla \mathcal{L}(\phi^{(t)})\|_{2}^{2} \\
\leq& \mathcal{L}(\phi^{(t)})-\mathcal{L}(\phi^{(t+1)}) + \frac{L \eta_{\phi}^{2}}{2}\|\zeta^{(t)}\|_{2}^{2}-(\eta_{\phi}-L\eta _{\phi}^{2})\langle\nabla \mathcal{L}(\phi^{(t)}), \zeta^{(t)}\rangle .
\end{aligned}
$$

\noindent Taking summation on both sides, we have \\
$$
\begin{aligned}
& \sum_{t=1}^{T}(\eta_{\phi}-\frac{L\eta_{\phi}^{2}}{2})\|\nabla \mathcal{L}(\phi^{(t)})\|_{2}^{2}  \\
\leq & \mathcal{L}(\phi^{(1)})-\mathcal{L}(\phi^{(T+1)}) + \frac{L \eta_{\phi}^{2}}{2}\sum_{t=1}^{T}\|\zeta^{(t)}\|_{2}^{2}-\sum_{t=1}^{T}(\eta_{\phi}-L\eta_{\phi}^{2})\langle\nabla \mathcal{L}(\phi^{(t)}), \zeta^{(t)}\rangle \\
& \leq \mathcal{L}(\phi^{(1)})-\mathcal{L}(\phi^{(T+1)}) +\frac{L \eta_{\phi}^{2} T \sigma^{2}}{2}
\end{aligned}
$$

\noindent Thus we have \\
$$
\begin{aligned}
\min _{t} \mathbb{E}\|\nabla \mathcal{L}(\phi^{(t)})\|^{2} 
\leq & \frac{\sum_{t=1}^{T}(\eta_{\phi}-\frac{L \eta_{\phi}^{2}}{2}) \|\nabla \mathcal{L}(\phi^{(t)})\|_{2}^{2}}{\sum_{t=1}^{T}(\eta_{\phi}-\frac{L \eta_{\phi}^{2}}{2})}\\
\leq & \frac{2 \mathcal{L}(\phi^{(1)}) - 2\mathcal{L}(\phi^{(T+1)})  +L \eta_{\phi}^{2} T \sigma^{2}}{\sum_{t=1}^{T}(2 \eta_{\phi}-L \eta_{\phi}^{2})}\\
\leq & \frac{2 \mathcal{L}(\phi^{(1)}) - 2\mathcal{L}(\phi^{(T+1)}) +L \eta_{\phi}^{2} T \sigma^{2}}{T \eta_{\phi}}\\
 =   & \frac{2 \mathcal{L}(\phi^{(1)}) -2\mathcal{L}(\phi^{(T+1)})}{T \eta_{\phi}}+L \sigma^{2} \eta_{\phi}\\
\leq & \frac{2 \mathcal{L}(\phi^{(1)})}{T} \max \{L, \frac{\sqrt{T}}{C}\}+\frac{2| \mathcal{L}(\phi^{(T+1)})|}{T}\max \{L, \frac{\sqrt{T}}{C}\} + L \sigma^{2} \min \{\frac{1}{L}, \frac{C}{\sqrt{T}}\} \\
= & \mathcal{O}(\frac{1}{\sqrt{T}}) .
\end{aligned}
$$
where we let $\eta_{\phi}=\min \{\frac{1}{L},\frac{C}{\sqrt{T}}\}$ for some $C > 0$.

\subsection{Proof of Theorem 2}
To prove Theorem 2, we first introduce the following three lemmas.
% \begin{lemma} \label{lemma1}
% Assume that $\| \mathbf{q}_i\|_{\infty} \leq E$ holds for all $x_i \in \mathcal{X}$, where $\mathbf{q}_i = f(g(x);\phi)$ is cluster assignment probability from the MLP classification model. We define the empirical risk and its expectation as
% To simplify the proof, we consider the case where the neighbors contain all training samples. Then the empirical and expected risk are defined as
\begin{lemma} \label{lemma1}
The empirical and expected risks of $\mathcal{L}_{I}(f(g(\mathcal{D});\phi))$ are defined as
$$
\begin{aligned}
\widehat{\mathcal{L}}_{n}(f) &=-\frac{1}{n} \sum_{i=1}^{n} \log \mathbf{q}_{i}^{T} \mathbf{q}_{i'},
\end{aligned}
$$

and
$$
\mathcal{L}(f) = -\mathbb{E} \left[  \log  \mathbf{q}^{T} \mathbf{q}_{i'} \right].
$$
where $\mathbf{q_i}=f(g(x_i);\phi)$. $x_{i'}$ is a randomly selected sample from the nearest neighbor set $\mathcal{N}_k(x_i)$, and $\mathbf{q}_{i'}$ is the soft cluster assignment of $x_{i'}$. 
With probability at least $1-\delta$, the following inequality holds\\
$$ 
\mathcal{L}(f) \le \widehat{\mathcal{L}}_{n}(f) + \frac{2\mu_{n}^{-1}}{\sqrt{n}}+(2+2k')\log\mu_{n}^{-1} \sqrt{\frac{\log \delta^{-1}}{2 n}}.
$$
\end{lemma} 
\emph{Proof.}
% $$
% \begin{aligned}
% \widehat{\mathcal{L}}_{n}(f) &=-\frac{1}{n} \sum_{i=1}^{n} \log \mathbf{q}_{i} \mathbf{q}_{i'}
% \end{aligned}
% $$ 
Let $\mathcal{D}'=(\mathcal{D}-{x_r})\cup x_{r'}$. The empirical risks on $\mathcal{D}'$ is denoted as  $\widehat{\mathcal{L}}_{n}'(f)$. According to Assumption~\ref{assump:num_neighbor}, let $\{x_{r1}, x_{r2}, \dots, x_{rl}\}$ be the set with $x_r$ as neighbor and $l\leq k'$. Then we have  
\begin{equation*}
    \begin{aligned}
\sup _{f \in \mathcal{F}}\left|\widehat{\mathcal{L}}_{n}(f)-\widehat{\mathcal{L}}_{n}^{'}(f)\right| \leq & \sup _{f \in \mathcal{F}}\frac{1}{n} \left(\left| (\log \mathbf{q}_r^{T} \mathbf{q}_{r'}- \log \bar{\mathbf{q}}_{r}^{T}\bar{\mathbf{q}}_{r'}) \right| + |\sum_{j=1}^{l}(\log \mathbf{q}_{rj}^T \mathbf{q}_{r}-\log \bar{\mathbf{q}}_{rj}^T \mathbf{q}_{rj'})|\right)\\
% \leq & \sup_{f \in \mathcal{F}} \frac{1}{n} \left| (\mathbf{q}_r \mathbf{q}_{r'}+\bar{\mathbf{q}}_r\bar{\mathbf{q}}_{r'}-2)\right| \\
\leq & \frac{(2+2k')\log \mu_n^{-1}}{n},
\end{aligned}
\end{equation*}
where the last inequality is according to Assumption~\ref{assump:knn}. \\

Let $\{\sigma_1,\sigma_2,\dots,\sigma_n\}$ be i.i.d. independent random variables taking values in $\{-1,1\}$ and $\bar{S} := \{\bar{x}_1,\dots,\bar{x}_n\}$ be independent of $S$, then we have
\begin{equation*}\small
\begin{aligned}
    & \mathbb{E}_{S} \left[ \mathop{\rm sup}_{f \in \mathcal{F}}|\mathcal{L}(f)-\widehat{\mathcal{L}}_{n}(f)| \right]\\
    % = & \mathbb{E}_{S} \left[\mathop{\rm sup}_{f_{\mathcal{Q}} \in \mathcal{F}} \frac{1}{n}\Bigg| \sum_{i=1}^n \log q_i q_{i'} - \mathbb{E}_{q,q'}[\log q q'] \Bigg|\right] \\
    = & \mathbb{E}_{S,\bar{S},\sigma} \left[\sup _{f \in \mathcal{F}} \frac{1}{n}\Bigg| \sum_{i=1}^n \sigma_{i}(\log \mathbf{q}_i^{T} \mathbf{q}_{i'} - \log \bar{\mathbf{q}}_i^{T} \bar{\mathbf{q}}_{i'}) \Bigg|\right] \\
    \leq & 2 \mathbb{E}_{S,\sigma} \left[\sup _{f \in \mathcal{F}} \frac{1}{n}\Bigg| \sum_{i=1}^n \sigma_{i}\log \mathbf{q}_i^{T} \mathbf{q}_{i'} \Bigg|\right] \\
    \leq & 2 \mathbb{E}_{S,\sigma} \left[\sup _{f \in \mathcal{F}} \frac{1}{n}\Bigg| \sum_{i=1}^n \sigma_{i}(\frac{1}{\mathbf{q}_i^{T} \mathbf{q}_{i'}}-1)  \Bigg|\right] \\    
    \leq & 2 \mathbb{E}_{S,\sigma} \left[\sup _{f \in \mathcal{F}} \frac{1}{n}\Bigg| \sum_{i=1}^n \sigma_{i}\frac{1}{\mathbf{q}_i^{T} \mathbf{q}_{i'} } \Bigg|\right] \\
    \leq & 2 \mathbb{E}_{S,\sigma} \left[\sup _{f \in \mathcal{F}} \frac{1}{n}\Bigg( \sum_{i=1}^n\left(\frac{1}{\mathbf{q}_i^{T} \mathbf{q}_{i'}}\right)^{2} \Bigg)^{\frac{1}{2}}\right] \\
    \leq&\frac{2\mu_{n}^{-1}}{\sqrt{n}}
\end{aligned}
\end{equation*}
where the second to last inequality is obtained by Khintchine-Kahane inequality~\cite{latala1994best} and the last inequality is obtained by Assumption~\ref{assump:knn}.

Thus according to the McDiarmid inequality~\cite{mohri2018foundations}, with probability at least $1-\delta$ for any $f\in \mathcal{F}$, we have
$$ 
\mathcal{L}(f) \le \widehat{\mathcal{L}}_{n}(f) + \frac{2\mu_{n}^{-1}}{\sqrt{n}}+(2+2k')\log\mu_{n}^{-1} \sqrt{\frac{\log \delta^{-1}}{2 n}}.
$$

% Let $\phi(h)=h^2$, $\| h\|_{\infty}=E$, $l$- $l=2E$

\begin{lemma}\label{lemma2}
We define the empirical its expectation risks of $\mathcal{L}_{B}(f(g(\mathcal{D});\phi))$ as 
\begin{equation*}\small
    \widehat{\mathcal{L}}_{n}(f) = -\sum_{l=1}^{c} \Bigg(\frac{1}{n} q_{il} \Bigg) \log \Bigg( \frac{1}{n}\sum_{i=1}^n q_{il} \Bigg),
\end{equation*}
and
\begin{equation*}\small
    \mathcal{L}(f) =  -\sum_{l=1}^c\mathbb{E}(q^{l})\log \mathbb{E}(q^{l}).
\end{equation*}
where $\mathbf{q_i}=f(g(x_i);\phi)$. 
With probability at least $1-\delta$, the following inequality holds
\begin{equation*}\small
\mathcal{L}(f) \leq \widehat{\mathcal{L}}_{n}(f) + \frac{2C}{\sqrt{n}} + C\sqrt{\frac{\log \delta^{-1}}{2n}}.
\end{equation*}
where $C=|\log\xi+1|$ is a bounded constant and $\xi$ is a constant according to the Lagrange Mean Theorem of the function $g(x) = x\log x$.
\end{lemma} 

\emph{Proof.} Let $S=\{x_1,\cdots,x_n\}$ and $S'=(S-{x_r})\cup \bar{x}_{r}$. The empirical risks on $S$ and $S'$ are denoted as $\widehat{\mathcal{L}}_{n}(f)$ and $\widehat{\mathcal{L}}_{n}'(f)$. Define $h(x)=x\log x$, according to the Lagrange Mean Theorem, there exists constant $\xi$ such that $|h(x)-h(y)| \leq |\log \xi + 1||x-y|$. We have
\begin{equation*}
    \begin{aligned}
&\sup _{f \in \mathcal{F}}\left|\widehat{\mathcal{L}}_{n}(f)-\widehat{\mathcal{L}}_{n}'(f)\right| \\
\leq & \sup _{f \in \mathcal{F}}\frac{1}{n}  \sum_{l}^c \left|\log\xi + 1\right| \left|q_{rl}-\bar{q}_{rl}\right|\\
\leq & \frac{\left| \log\xi + 1 \right|}{n} \\
= & \frac{C}{n}. 
\end{aligned}
\end{equation*}
where $C = \left| \log\xi + 1 \right|$. \\
Let $\sigma_1,\sigma_2,\dots,\sigma_n$ be i.i.d. independent random variables taking values in $\{-1,1\}$ and $\bar{S} := \{\bar{x}_1,\dots,\bar{x}_n\}$ be the independent copy of $S := \{x_1, \dots, x_n\}$. Then we have
\begin{equation*}\small
\begin{aligned}
    & \mathbb{E}_{S} \left[ \mathop{\rm sup}_{f_{\mathcal{Q}} \in \mathcal{F}}|\mathcal{L}(f)-\widehat{\mathcal{L}}_{n}(f)| \right]\\
    \leq & \mathbb{E}_{S,\bar{S},\sigma} \left[\sup _{f_{\mathcal{Q}} \in \mathcal{F}} \frac{1}{n}\sum_{i=1}^n\sigma_i \left(\sum_{l=1}^{c} |\log\xi + 1| | q_{il}-\bar{q}_{il} |\right)  \right] \\  
    \leq & 2|\log\xi + 1|\mathbb{E}_{S,\sigma} \left[\sup _{f_{\mathcal{Q}} \in \mathcal{F}} \frac{1}{n}\sum_{i=1}^n\sigma_i \sum_{l=1}^{c} q_{il}  \right] \\ 
    \leq & 2|\log\xi + 1|\mathbb{E}_{S,\sigma} \left[\sup _{f_{\mathcal{Q}} \in \mathcal{F}} \frac{1}{n}\left(\sum_{i=1}^n \left[\sum_{l=1}^{c} q_{il}\right]^2 \right)^{\frac{1}{2}} \right] \\
    \leq & \frac{2|\log\xi + 1|}{\sqrt{n}} \\
    = & \frac{2C}{\sqrt{n}}.  \\
\end{aligned}
\end{equation*}
where $C = \left| \log\xi + 1 \right|$. \\
Thus according to the McDiarmid inequality~\cite{mohri2018foundations}, with probability at least $1-\delta$ for any $f\in \mathcal{F}$, we have
$$ 
\mathcal{L}(f) \leq \widehat{\mathcal{L}}_{n}(f) + \frac{2C}{\sqrt{n}} + C\sqrt{\frac{\log \delta^{-1}}{2n}}.
$$
\begin{lemma}\label{lemma3}
We define the empirical and expectation risks of $\mathcal{L}_{IS}(f(g(\mathcal{D});\phi))$ as
\begin{equation*}\small
    \widehat{\mathcal{L}}_{n}(f) = -\frac{1}{n}\sum_{i=1}^{n}\sum_{l=1}^c p_{il}\log q_{il},
\end{equation*}
and
\begin{equation*}\small
    \mathcal{L}(f) = -\sum_{l=1}^c \mathbb{E}(p^{l}\log q^{l}) .
\end{equation*}
where $\mathbf{q}_i = f(g(x_i);\phi)$ and $\mathbf{p}_i$ is a one-hot pseudo-label.
With probability at least $1-\delta$, the following inequality holds
$$ 
\mathcal{L}(f) \le \widehat{\mathcal{L}}_{n}(f) + \frac{2\log\mu_p^{-1}}{\sqrt{n}} + 2\log\mu_p^{-1}\sqrt{\frac{\log \delta^{-1}}{2 n}}.
$$
\end{lemma} 
\emph{Proof.} Let $S=\{x_1,\cdots,x_n\}$ and $S'=(S-{x_r})\cup \bar{x}_{r}$. The empirical risks on $S$ and $S'$ are denoted as $\widehat{\mathcal{L}}_{n}(f)$ and $\widehat{\mathcal{L}}_{n}'(f)$. We have
\begin{equation*}
    \begin{aligned}
& \sup _{f \in \mathcal{F}}\left|\widehat{\mathcal{L}}_{n}(f)-\widehat{\mathcal{L}}_{n}'(f)\right| \\
\leq & \sup _{f \in \mathcal{F}}\frac{1}{n} \left| \sum_{l}^c (p_{rl}\log q_{rl}- \bar{p}_{rl}\log \bar{q}_{rl}) \right| \\
% \leq & \sup _{f \in \mathcal{F}}\frac{1}{n} \left| \sum_{l}^c (p_{rl}(\frac{1}{q_{rl}}-1) + \bar{p}_{rl}(\frac{1}{\bar{q}_{rl}}-1)) \right| \\
% \leq & \sup _{f \in \mathcal{F}}\frac{1}{n} \left| \sum_{l}^c (p_{rl}(q_{rl}-1) + \bar{p}_{rl}(\bar{q}_{rl}-1)) \right| \\
\leq & \frac{2c\log\mu_{p}^{-1}}{n}.
\end{aligned}
\end{equation*}
Next we analyze the upper bound of the expectation term, \emph{i.e.}, $\mathbb{E}_{S} \left[\mathop{\rm sup}_{f \in \mathcal{F}}|\mathcal{L}(f)-\widehat{\mathcal{L}}_{S}(f)|\right ]$. Let $\sigma_1,\sigma_2,\dots,\sigma_n$ be i.i.d. independent random variables taking values in $\{-1,1\}$ and $\bar{S} := \{\bar{x}_1,\dots,\bar{x}_n\}$ be the independent copy of $S := \{x_1, \dots, x_n\}$. Then we have
\begin{equation*}\small
\begin{aligned}
    & \mathbb{E}_{S} \left[ \mathop{\rm sup}_{f_{\mathcal{Q}} \in \mathcal{F}}|\mathcal{L}(f)-\widehat{\mathcal{L}}_{n}(f)| \right]\\
    = & \mathbb{E}_{S,\bar{S},\sigma} \left[\sup _{f_{\mathcal{Q}} \in \mathcal{F}} \frac{1}{n}\Bigg| \sum_{i=1}^n \sigma_{i}(\sum_{l=1}^{c}(p_{il}\log q_{il} - \bar{p}_{il}\log \bar{q}_{il}) ) \Bigg|\right] \\
    \leq & 2 \mathbb{E}_{S,\sigma} \left[\sup _{f_{\mathcal{Q}} \in \mathcal{F}} \frac{1}{n}\Bigg| \sum_{i=1}^n \sigma_{i} \sum_{l=1}^{c} p_{il}\log q_{il}  \Bigg|\right] \\
    % \leq & 2 \mathbb{E}_{S,\sigma} \left[\sup _{f_{\mathcal{Q}} \in \mathcal{F}} \frac{1}{n}\Bigg| \sum_{i=1}^n \sigma_{i}\sum_{l=1}^{c} p_{il} (\frac{1}{q_{il}}-1)  \Bigg|\right] \\ 
    \leq & 2 \mathbb{E}_{S,\sigma} \left[\sup _{f_{\mathcal{Q}} \in \mathcal{F}} \frac{1}{n}\left( \sum_{i=1}^n \left[\sum_{l=1}^{c}p_{il} \log q_{il}\right]^2 \right)^{\frac{1}{2}}\right] \\
    \leq & \frac{2c\log\mu_{p}^{-1}}{\sqrt{n}}.
\end{aligned}
\end{equation*}
Thus according to the McDiarmid inequality~\cite{mohri2018foundations}, with probability at least $1-\delta$ for any $f\in \mathcal{F}$, we have
$$ 
\mathcal{L}(f) \le \widehat{\mathcal{L}}_{n}(f) + \frac{2c\log\mu_p^{-1}}{\sqrt{n}} + 2c\log\mu_p^{-1}\sqrt{\frac{\log \delta^{-1}}{2 n}}.
$$
Now we give the proof of Theorem 2. \\
\emph{Proof.} We define the empirical and expectation risks of $\mathcal{L}(f)$ in Eq. (\ref{loss_overall}) as
\begin{equation*}\small
\begin{aligned}
    \widehat{\mathcal{L}}_{n}(f(g(\mathcal{D});\phi)) = 
    & -\frac{1}{n} \sum_{i=1}^{n} \log \mathbf{q}_{i}^{T} \mathbf{q}_{i'}  -\beta\frac{1}{n}\sum_{i=1}^n\sum_{l=1}^c p_{il}\log q_{il}- \lambda\sum_{l=1}^c \Bigg( \frac{1}{n}\sum_{i=1}^n q_{il} \Bigg) \log \Bigg( \frac{1}{n}\sum_{i=1}^n q_{il} \Bigg)
\end{aligned}
\end{equation*}
and
\begin{equation*}\small
\begin{aligned}
    \mathcal{L}(f(g(\mathcal{X});\phi)) = 
    & -\mathbb{E} \left[  \log  \mathbf{q}^{T} \mathbf{q}_{rs} \right] - \beta\sum_{l=1}^c  \mathbb{E}(p^{l}\log q^{l})-\lambda \sum_{l=1}^c\mathbb{E}(q^{l}\log q^{l}) .
\end{aligned}
\end{equation*}
According to Lemma \ref{lemma1}, \ref{lemma2} and \ref{lemma3}, with probability at least $1-\delta$ for any $f\in \mathcal{F}$, we have
\begin{equation*}
    \mathcal{L}(f(g(\mathcal{X});\phi)) \leq \widehat{\mathcal{L}}_{n}(f(g(\mathcal{D});\phi)) + \frac{\tilde{c}_1}{\sqrt{n}} + \tilde{c}_2\sqrt{\frac{1}{2n}\log \delta^{-1}}.
\end{equation*}
where $\tilde{c}_1=2\mu_{n}^{-1}+2C\beta+2c\lambda\log\mu_p^{-1}$ and $\tilde{c}_2= (2+2k')\log\mu_{n}^{-1} + C\beta +2c\lambda\log\mu_p^{-1}$. $C$ is a constant for the function $x\log x$. This finishes the proof.

\end{document}